\documentclass[final,12pt,3p]{elsarticle}

\usepackage[colorlinks=true]{hyperref}
\usepackage{graphicx}
\usepackage{subcaption}

\usepackage{amssymb}

\usepackage{lipsum}
\usepackage[labelsep=period,singlelinecheck=off,textformat=period]{caption}
\usepackage{subcaption}
\usepackage{fancyhdr} 
\usepackage{epsfig}
\usepackage{timesmt}
\usepackage{mathtools,nccmath}
\usepackage{mathrsfs}
\usepackage{wrapfig}
\usepackage[noabbrev]{cleveref}
\creflabelformat{equation}{#2\textup{#1}#3}

\usepackage{pdflscape}
\usepackage{afterpage}
\usepackage{float}
\usepackage{url}
\usepackage{enumitem}
\usepackage{accents}
\usepackage{cuted}
\usepackage{dirtytalk}
\usepackage{multirow}
\usepackage{multicol}
\usepackage{booktabs}
\usepackage{array}
\usepackage{makecell}
\usepackage{colortbl}
\usepackage{soul}
\newcolumntype{n}{>{\columncolor{blue!5}}c}

\usepackage{pbox}
\usepackage[ruled,commentsnumbered,linesnumbered,boxed]{algorithm2e}
\SetArgSty{textnormal}
\SetKwComment{Comment}{/* }{ */}

\usepackage[flushleft]{threeparttable}
\usepackage{algpseudocode}
\usepackage{setspace}
\usepackage{courier}
\usepackage{psfrag,pstool}

\newcommand{\comment}[1]{}

\Crefname{equation}{Eq.}{Eqs.}
\Crefname{figure}{Fig.}{Figs.}
\Crefname{tabular}{Tab.}{Tabs.}
\crefname{algocf}{alg.}{algs.}
\Crefname{algocf}{Algorithm}{Algorithms}
\Crefname{fct}{Fact}{Facts }

\usepackage{amsmath}
\usepackage{amssymb}
\usepackage{amsthm}
\usepackage{mathrsfs}

\usepackage{bm}
\usepackage{mathtools}
\usepackage{nicefrac}

\theoremstyle{plain}

\theoremstyle{definition}
\newtheorem{defn}{Definition}[section]

\theoremstyle{remark}
\newtheorem{rem}{Remark}

\newcommand{\transpose}{^{\mathrm{T}}}

\renewcommand{\Re}{\mathbb{R}}

\newcommand{\x}{{\mathbf x }}

\newcommand{\X}{{\mathbf X}}

\newcommand{\Y}{{\mathbf Y}}

\newcommand{\Z}{{\mathbf Z}}

\newcommand{\thetaa}{\bm{\theta}}

\newcommand{\ie}{i.e., }
\newcommand{\eg}{e.g., }
\newcommand{\supth}[1]{\ensuremath{{#1}^{\text{th}}}}

\usepackage{pgfplots}
\pgfplotsset{compat=1.16}
\usepackage{tikz}

\usetikzlibrary{shapes.misc}
\usetikzlibrary{arrows,arrows.meta,calc,backgrounds}

\tikzstyle{block} = [draw,rectangle,thick,minimum height=2em,minimum width=2em]
\tikzstyle{sum} = [draw,circle,inner sep=0mm,minimum size=2mm]
\tikzstyle{connector} = [->,thick]
\tikzstyle{line} = [thick]
\tikzstyle{branch} = [circle,inner sep=0pt,minimum size=1mm,fill=black,draw=black]
\tikzstyle{guide} = []
\tikzset{>=latex}

\makeatletter
\newcommand{\hathat}[1]{%
	\begingroup%
	\let\macc@kerna\z@%
	\let\macc@kernb\z@%
	\let\macc@nucleus\@empty%
	\hat{\raisebox{.35ex}{\vphantom{\ensuremath{#1}}}\smash{\hat{#1}}}%
	\endgroup%
}
\makeatother

\creflabelformat{equation}{#2(#1)#3}
\crefrangelabelformat{equation}{#3(#1)#4 to #5(#2)#6}

\labelcrefformat{subequation}{#2(#1)#3}
\labelcrefrangeformat{subequation}{#3(#1)#4 to #5(#2)#6}

\renewcommand{\mathbf}[1]{\bm{#1}}

\graphicspath{{./tikzpics/}{./figures/}}

\usepackage{lineno}

\journal{xxx}

\begin{document}
	
\begin{frontmatter}
		

		
\title{Time-dependent density estimation using binary classifiers}
		
		
\author[label1]{Agnimitra Dasgupta}
\author[label1]{Javier Murgoitio-Esandi}
\author[label1]{Ali Fardisi}
\author[label1]{Assad A Oberai\corref{cor1}}

\cortext[cor1]{Corresponding author}
\ead{aoberai@usc.edu}
		
\affiliation[label1]{organization={{Department of Aerospace \& Mechanical Engineering, University of Southern California}},
city={Los Angeles},
postcode={90089}, 
state={California},
country={USA}}
		
\begin{abstract}
We propose a data-driven method to learn the time-dependent probability density of a multivariate stochastic process from sample paths, assuming that the initial probability density is known and can be evaluated. Our method uses a novel time-dependent binary classifier 
trained using a contrastive estimation-based objective that trains the classifier to discriminate between realizations of the stochastic process at two nearby time instants. Significantly, the proposed method explicitly models the time-dependent probability distribution, which means that it is possible to obtain the value of the probability density within the time horizon of interest. Additionally, the input before the final activation in the time-dependent classifier is a second-order approximation to the partial derivative, with respect to time, of the logarithm of the density. We apply the proposed approach to approximate the time-dependent probability density functions for systems driven by stochastic excitations. We also use the proposed approach to synthesize new samples of a random vector from a given set of its realizations. In such applications, we generate sample paths necessary for training using stochastic interpolants. Subsequently, new samples are generated using gradient-based Markov chain Monte Carlo methods because automatic differentiation can efficiently provide the necessary gradient. Further, we demonstrate the utility of an explicit approximation to the time-dependent probability density function through applications in unsupervised outlier detection. Through several numerical experiments, we show that the proposed method accurately reconstructs complex time-dependent,  multi-modal, and near-degenerate densities, scales effectively to moderately high-dimensional problems, and reliably detects rare events among real-world data. 
\end{abstract}
		
		
		
\begin{keyword}
   Probabilistic learning, density estimation, generative modeling, data-driven modeling, MCMC generators
\end{keyword}		
\end{frontmatter}
	
\section{Introduction}\label{sec:introduction}

A problem that routinely arises in scientific and statistical computing involves approximating an evolving probability density function from a finite number of sample paths of the underlying multivariate stochastic process. In computational physics, for example, the approximation of time-dependent densities from sample paths of stochastic processes associated with stochastic differential equations (SDEs) with a given or terminal initial condition is useful for solving partial differential equations (PDEs)~\cite{han2018solving,cho2016numerical}. Density estimation also has applications in  uncertainty quantification~\cite{soize2023overview}, surrogate modeling~\cite{soize2024probabilistic,zhong2023surrogate}, Bayesian inference~\cite{el2012bayesian,dasgupta2025conditional}, and rare events simulation~\cite{dasgupta2024rein}. In machine learning too, density estimation has many applications ranging from generative modeling to out-of-distribution detection or anomaly~\cite{murphy2022probabilistic,murphy2023probabilistic,sugiyama2012density}. 

Several methods have been proposed to estimate time-dependent densities from sample paths. The spectrum of these methods ranges from indirect to direct approaches. Indirect methods provide samples or sample paths from which the density may be estimated. For instance, a class of indirect methods attempt to identify the underlying SDE or the deterministic velocity field that describes the evolution of the particles realized from the stochastic process, ultimately learning to simulate new sample paths~\cite{garcia2017nonparametric,ruttor2013approximate,gonzalez1998identification,brunton2016discovering,riseth2017operator,dai2020detecting,yang2022generative}. We classify these approaches as indirect because the density must be estimated using kernel-based methods~\cite{wasserman2006all} from path simulations or using appropriate change of variables formulae~\cite{kothe2023review}. Yet another class of indirect methods uses deep generative models, such as generative adversarial networks, to synthesize new sample paths~\cite{yang2020physics,yang2022generative} and then learn the probability density. 

In contrast to the indirect methods, direct methods explicitly approximate the time-dependent density. For instance, when sample paths of an SDE with known initial or terminal conditions are observed, neural networks can be used to directly approximate or estimate the time-dependent probability density from sample paths using the reconstruction error of the known initial or terminal condition \cite{han2018solving,guler2019towards,beck2021solving} as the loss function. In some cases, this loss is supplemented by a physics-informed loss~\cite{chen2021solving}. However, these approaches often require knowledge of the underlying SDE, which may not be available in practical applications. More recently, \citet{lu2022learning} proposed a time-dependent normalizing flow to approximate evolving probability densities from sample paths in a purely \emph{data-driven} paradigm without recourse to identifying SDEs or solving Fokker-Planck equations. However, normalizing flows use special invertible architectures that have large memory footprints~\cite{dasgupta2024dimension}, which hinders their scalability to high-dimensional problems. Alternatively, the time-dependent density estimation tasks may be divided into smaller, manageable tasks of density-ratio estimation between two arbitrary densities at successive time steps, where observations are available. Density-ratio estimation methods~\cite{gutmann2010noise,rhodes2020telescoping,sugiyama2012density} can then be utilized to directly estimate the density along the stochastic process. However, it is unclear how the density can be evaluated between the time points where data is available. Therefore, the density-ratio estimation techniques are not useful for building continuous approximations to the time-dependent probability density.  


Nestled between the direct and indirect approaches on either end of the spectrum are approaches that are primarily designed to generate sample paths of the underlying stochastic process, but can be extended to estimate the associated density. Dynamical approaches to measure transport that transform samples between successive time instants along a stochastic process fall under this category, where the core idea is to learn an ODE or SDE that can achieve the desired transformation~\cite{tsimpos2025optimal}. Flow matching~\cite{lipman2022flow} methods --- such as continuous normalizing flows~\cite{chen2018neural,grathwohl2018ffjord,onken2021ot}, dual diffusion models~\cite{su2022dual}, stochastic interpolants~\cite{albergo2022building,albergo2023stochastic,liu2022flow,liu2022rectified} and Schrodinger Bridges~\cite{pavon2021data,de2021diffusion,chen2021likelihood,pooladian2024plug} --- fall within this category. However, there are two potential drawbacks to using such methods for time-dependent density estimation. First, evolving samples over the entire process will require the construction of multiple transport maps, one each between successive time steps. Second, log-density evaluation in flow matching models typically involves expensive path-wise integration of reverse-time ODEs or SDEs, which can make density estimation over large time horizons computationally intractable. 

In this work, we propose a direct, data-driven, and scalable approach to time-dependent density estimation. Our approach is motivated by the ability of an optimally trained discriminative classifier to estimate the ratio between two probability densities~\cite{gutmann2010noise,duvenaud2020your}. We build on this insight, making a time-dependent binary classifier the core of our approach. This classifier is trained to discriminate between samples drawn from two distributions that are nearby in time (separated by the interval $\Delta t$).  We implement a special architecture for the classifier that accomplishes three goals. First, it includes time as an explicit input to the classifier so that the learned weights are shared across all time instances of the stochastic process. Second, within the classifier, we explicitly identify a network whose output approximates the rate of change of the log-density of the target probability density function. Thereby lending a degree of interpretability to our method. Finally, we incorporate the time interval ($\Delta t$) between the two nearby densities into the classifier in a special way that ensures numerical stability as this interval becomes small. Once trained, the classifier can be used to evaluate the density at any point in the data space and at any time, by evaluating a simple sum that approximates an integral in time. This is in contrast to other methods that need to evaluate a path in (data) space-time and then an integral along that path. 
This path-independent formulation, combined with the architectural flexibility offered by a classifier free from invertibility constraints, yields a method that is both memory-efficient and scalable.

\paragraph{Summary of contributions} We briefly summarize our contributions as follows:
\begin{enumerate}[itemsep=1pt,leftmargin=*]
    \item We introduce a novel data-driven approach to explicitly approximate the temporal evolution of probability densities without prior knowledge of the underlying stochastic dynamics. This is possible using a novel time-dependent classifier that learns the partial time derivative of the log-density from samples drawn at different time instances. 
    \item We propose a special architecture for the classifier and leverage a contrastive estimation-based loss function to train the time-dependent classifier from sample data. The special architecture and the stationary point of the loss function help lend an interpretation to the output of the neural network within the classifier as a second-order accurate approximation of the partial time derivative of the log-density. It also ensures that the proposed method is stable as the time interval between the contrastive densities becomes small. 
    \item Using stochastic interpolants, we perform classic or static density estimation with the proposed approach. The stochastic interpolant helps induce a synthetic process bridging a tractable latent distribution with the data-generating distribution. Further, the learned density permits efficient gradient-based sampling via automatic differentiation. 
    \item We demonstrate that the method accurately captures time-dependent densities of stochastic systems driven by white noise, and estimates densities that have disconnected supports, are multimodal, or are supported over manifolds from samples.
    \item The method also achieves competitive or superior results on benchmark rare event detection tasks compared to state-of-the-art methods.
\end{enumerate}

\paragraph{Organization of the paper} The remainder of the paper is organized as follows. \Cref{sec:time-dependent-density-estimation} sets up the main problem we wish to tackle, introduces the proposed approach, and provides relevant theory. \Cref{sec:generative-model} shows how the proposed method can be used specifically for time-independent density estimation from a given sample set, and subsequent sample generation. \Cref{sec:numerical_examples} provides numerical examples where the proposed approach is used for time-dependent density estimation for dynamical systems driven by stochastic excitations (\Cref{subsec:results-javier}), time-independent density estimation and sample generation (\Cref{sec:results-dens-est-sampling}), and rare event detection (\Cref{subsec:outlier-detection-results}). Finally, \Cref{sec:conclusion} concludes the paper. 

\section{Time-dependent density estimation}\label{sec:time-dependent-density-estimation}

\subsection{Problem setup}\label{subsec:problem-setup}

The following is necessary to set up the problem we wish to solve and clarify the notation we will use throughout this paper. 
\begin{itemize}
    \item[\textit{(i)}] \textit{Notation:} Let $\left\{ \X_t , t \in [0, T ]\right\} $ denote a $\Re^{n}$-valued stochastic process indexed by $t \in [0, T]$. Therefore, $\X_t$ is a random vector with values in $\Re^n$ for which the probability distribution is defined by the probability density function $\rho_t(\x)$. We will use $\x_t^{(i)}$ to denote the \supth{i} realization of $\X_t$. So, $\{ \x_t^{(1)}, \x_t^{(2)}, \ldots, \x_t^{(N)} \}$ denotes a sample of $N$ independent and identically distributed (iid) realizations of $\X_t$ drawn from $\rho_t$. 
    
    \item[\textit{(ii)}] \textit{Dataset:} The available dataset $\mathcal{D}$ consists of realizations of $\X_t$ observed at various time instants $t \in [0, T]$, \ie $\mathcal{D} = \left\{ \left\{   \x_{t_j}^{(i)}  , t_j \right\}_{i=1}^{N_{j}}\right\}_{j=1}^{N}$. Thus, the training data $\mathcal{D}$ consists of $N = N_{j} \times N_t$ observations of the stochastic process: $N_{t_j}$ observations of $\X_{t_j}$ at $N$ distinct time instants, where $t_0 = 0$ and $t_{N} = T$. We note that generally, any given path may not be observed at all different instants of time $t_j \in \{t_0, t_1, \ldots t_{N}\} $. Therefore, the available dataset corresponds to the case of `\emph{unpaired}' observations~\cite{yang2022generative}. However, the proposed method does not make any assumptions about the nature of the dataset and can work with both paired and unpaired datasets.
    \item[\textit{(iii)}] \textit{Assumptions:} We also assume that the initial density $\rho_0$ is known and analytically tractable \ie we assume that it is possible evaluate $\rho_0(\x) \;\; \forall \; \x \in \Re^n$. This assumption can be relaxed as long as $\rho_t$ is known and can be evaluated analytically at some point $t \in [0, T]$; see Remark~\ref{rem1:assumption} later.  Additionally, without a loss in generality, we assume that $\rho_t$-s are infinitely supported over $\Re^n$ for all $t \in [0, T]$. 
	\item[\textit{(iv)}] \textit{Objective:} The objective of this paper is to estimate $\rho_t \;\; \forall \; t \in [0, T]$ using the training dataset $\mathcal{D}$. 
\end{itemize}

\subsection{Proposed method}\label{subsec:method}

\subsubsection{Overview}

\begin{figure}[t]
    \centering
    \includegraphics[width=\linewidth]{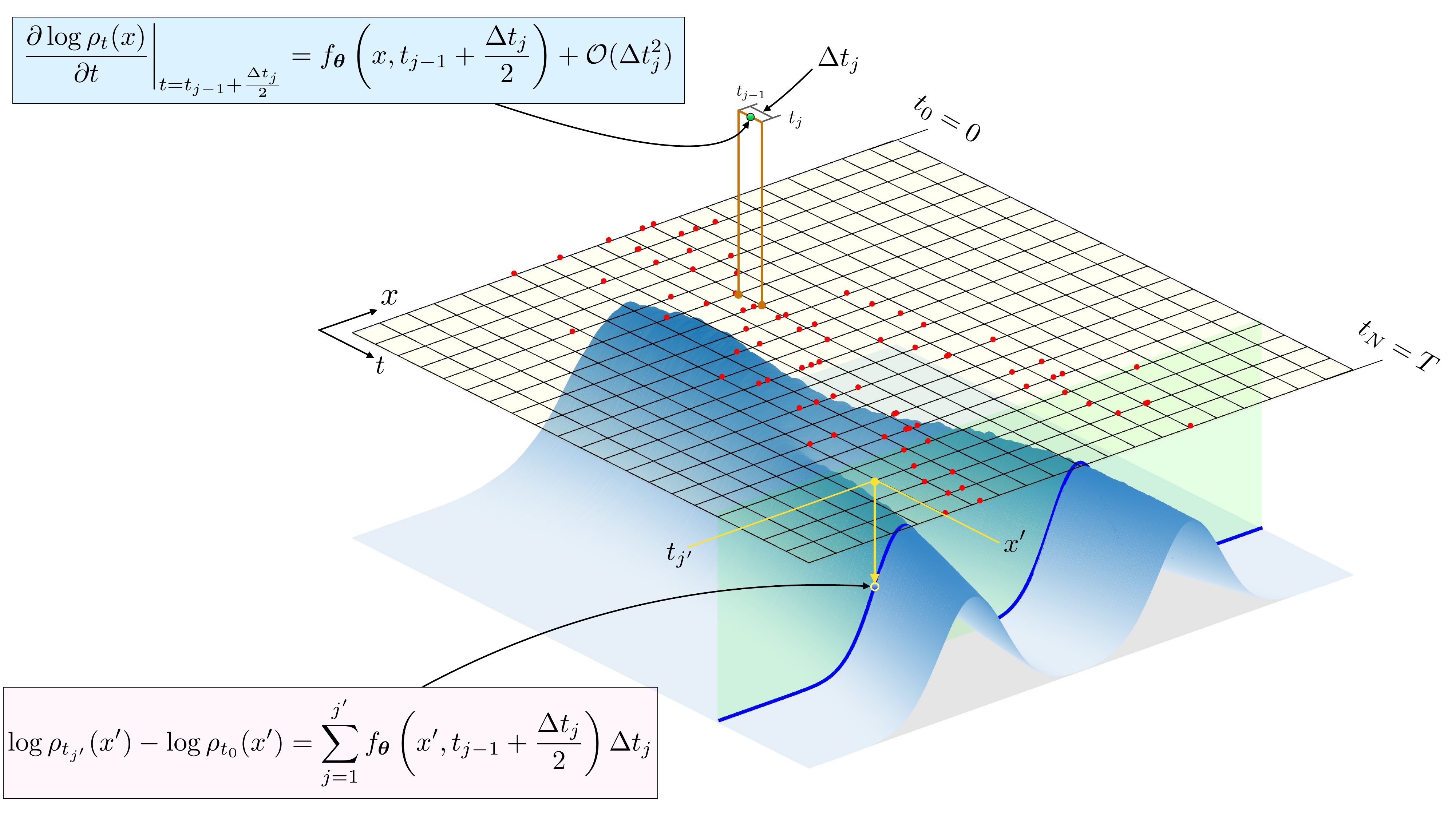}
    \caption{Overview of the proposed approach for time-dependent density estimation. $f_{\thetaa}$ learns the partial derivative of the log-density with respect to time from observations of the stochastic process (shown in red dots {\color{red}$\bullet$}). Subsequently, the log-density can be estimated via simple summation}
    \label{fig:overview}
\end{figure}

We propose to train a time-dependent classifier, which is built using the time-dependent network $f_{\thetaa}$, to approximate the partial derivative of the log-density $\log \rho_t$ with respect to time $t$ between successive time instants $t_{j-1}$ and $t_{j}$ where observations $\x_{t_{j-1}}^{(i)}$ and $\x_{t_{j}}^{(i)}$ of the stochastic process are available. To be precise, we will show that the time-dependent network $f_{\thetaa}$ approximates $\partial \log \rho_t(x)/ {\partial t}$ such that
\begin{equation}
    \frac{\partial \log \rho_t(\x)}{\partial t} \bigg\rvert_{t = t_{j-1} + \frac{\Delta t_j}{2}} = f_{\thetaa} \left(\x, t_{j-1} + \frac{\Delta t_j}{2} \right) + \mathcal{O}(\Delta t_j^2), 
\end{equation}
where $\Delta t_j = t_{j} - t_{j-1}$. Using the trained network, the log-density of the stochastic process can be evaluated for any $\x^\prime$ and ${t_{j^\prime}}$ simply using
\begin{equation}
    \log \rho_{t_{j^\prime}}(\x^\prime) - \log \rho_{0}(\x^\prime) = \sum_{j=1}^{j^\prime} f_{\thetaa} \left(\x^\prime, t_{j-1} + \frac{\Delta t_j}{2} \right) \Delta t_j ,
\end{equation}
which avoids any path-wise integration. \Cref{fig:overview} provides an overview of the proposed approach. We describe the classifier in \Cref{subsubsec:classifier}, develop the loss function used to train the classifier in \Cref{subsubsec:loss}, discuss the estimation of the log-density in \Cref{subsubsec:estimating-density}, and provide an interpretation for the classifier in \Cref{subsubsec:interpretation}. 

\subsubsection{Time-dependent classifier}\label{subsubsec:classifier}
Let $d_{\thetaa}(\x, t, \Delta t): \Re^n \times [0, T] \times [0, \epsilon] \to [0, 1]$ denote a time-dependent classifier with learnable parameters $\thetaa$ that accepts as input $\x$ along with a time instant $t \in (0, T)$ and a time gap $\Delta t$ such that $\epsilon = \max_{j} \Delta t_j$,  where $\Delta t_j = t_{j+1} - t_j$ and $j \in \{0, 1, \ldots , N_t - 1 \}$. We model $d_{\thetaa}$ as follows:
\begin{equation}\label{eq:time-dependent-classifier}
	d_{\thetaa}(\x, t, \Delta t) = \sigma\big( f_{\thetaa}(\x, t) \Delta t \big) =   \frac{1}{1 + \exp\left(- f_{\thetaa}(\x, t) \Delta t \right)}  ,
\end{equation}
where $f_{\thetaa}(\x, t)$ is a feed forward neural network that explicitly depends on time through the input $t$, and $\sigma(\cdot)$ denotes the sigmoid activation commonly used in binary classifiers. Note how $\Delta t$ scales the network's output before the sigmoid activation. The reasons for this particular choice of the classifier $d_{\thetaa}$ will become apparent in \Cref{subsubsec:loss}. They include ensuring that the output of $f_{\thetaa}$ does not vanish as $\Delta t \to 0$. Intuitively, however, distinguishing samples from $\X_{t}$ and $\X_{t + \Delta t}$ should be difficult when $\Delta t$ is small, so scaling the output of $f_{\thetaa}$ by $\Delta t$ helps ensure that $d_{\thetaa} \to 0.5$ as $\Delta t \to 0$. We also note that
\begin{equation}\label{eq:sigmoid-reciprocal}
	\log \left(  \frac{d_{\thetaa}(\x,t, \Delta t)}{1 - d_{\thetaa}(\x, t, \Delta t)} \right) = f_{\thetaa}(\x, t) \Delta t ,
\end{equation}
where $\log(\cdot)$ denotes the natural logarithm. \Cref{eq:sigmoid-reciprocal} will be useful later. 

\subsubsection{Loss function for training the classifier}\label{subsubsec:loss}
We propose to minimize the loss function
\begin{equation}\label{eq:loss-general}
    \mathsf{J}(\thetaa) = \sum_{j=1}^{N}  \Bigg\{ \int \mathsf{C}\left[ d_{\thetaa}\left(\x, \bar{t}_j , \Delta t_j \right), 0 \right] \rho_{t_{j-1}}(\x)   \mathrm{d}\x + \int\mathsf{C}\left[ d_{\thetaa}\left(\x, \bar{t}_j , \Delta t_j \right), 1 \right] \rho_{t_{j}}(\x) \mathrm{d}\x  \Bigg\}  
\end{equation}
to train the classifier $d_{\thetaa}$, where $\bar{t}_j = (t_{j-1} + t_{j}) /2 = t_{j-1} + \Delta t_j / 2$ is the mid point between successive time instants $t_{j-1}$ and $t_{j}$, and $\Delta t_j = t_j - t_{j-1}$.  In \Cref{eq:loss-general}, $\mathsf{C}: [0, 1] \times \{0,1\} \to \mathbb{A} $ is an appropriate cost function for classification and $ \mathrm{d}\x  = \mathrm{d}x_1 \, \mathrm{d}x_2 \, \ldots \mathrm{d}x_n$ is the Lebesgue measure over $\Re^n$. Possible candidates for $\mathsf{C}$ include the Brier score
\begin{equation}
	\mathsf{C}\left[a, b \right] = (b - a)^2 \in [0,1] := \mathbb{A},
\end{equation}
or the logarithmic score 
\begin{equation}
	\mathsf{C}\left[ a, b \right] = - b \log a -  (1 - b) \log (1 - a) \in [0, \infty) := \mathbb{A}.
\end{equation}
These cost functions are routinely used for binary classification~\cite{ferrer2022analysis}; the Brier and logarithmic scores are connected to the mean-squared-error and cross-entropy losses, respectively. We use the Brier score in this work, and the corresponding loss function $\mathsf{J}$ is
\begin{equation}\label{eq:loss}
\mathsf{J}(\thetaa) = \sum_{j=1}^{N}  \Bigg\{ \int \left[ d_{\thetaa}\left(\x, \bar{t}_j , \Delta t_j \right) \right]^2\rho_{t_{j-1}}(\x)   \mathrm{d}\x  + \int \left[ 1 - d_{\thetaa}\left(\x, \bar{t}_j, \Delta t_j \right) \right]^2 \rho_{t_{j}}(\x) \mathrm{d}\x  \Bigg\} . 
\end{equation}
Moreover, in practice, we approximate \Cref{eq:loss} with Monte Carlo samples from the dataset $\mathcal{D}$ as follows
\begin{equation}\label{eq:loss-MCS}
    \mathsf{J}(\thetaa) \approx \frac{1}{2N_{\mathrm{b}}} \sum_{j=1}^{N} \sum_{i=0}^{N_{\mathrm{b}}} \left\{ \left[ d_{\thetaa}\left(\x^{(i)}_{t_{j-1}}, \bar{t}_j, \Delta t_j\right)\right]^2 + \left[ 1 -  d_{\thetaa}\left(\x^{(i)}_{{t_{j}}},  \bar{t}_j, \Delta t_j\right)\right]^2 \right\} , 
\end{equation}
where $\bar{t}_j = (t_{j-1} + t_{j})/2$, $\Delta t_j = t_{j} - t_{j-1}$, and $\x^{(i)}_{t_j}$ are sampled uniformly from the dataset $\mathcal{D}$ at $t_j$. In \Cref{eq:loss-MCS}, $2N_{\mathrm{b}}$ is the effective batch size. We randomly replicate the observations within each batch if $N_b > N_{j}$ for some time instant $t_j$. Also, we minimize \Cref{eq:loss-MCS} using a suitable gradient-descent optimization algorithm~\cite{ruder2016overview}.

\begin{rem}[Relation to Noise Contrastive Estimation]
    The loss function we propose is a time-dependent adaptation of Noise Contrastive Estimation (NCE)~\cite{gutmann2010noise} --- a method for estimating the unnormalized probability densities by converting the task into a binary classification problem. Specifically, a classifier is trained to distinguish between samples from a target (data) distribution and a known noise distribution. The optimal classifier recovers the density ratio between the two, and since the noise distribution is known, this enables estimation of the data distribution up to a normalizing constant. Our method adopts a similar philosophy but in a dynamical setting. Instead of contrasting static data and noise distributions, we consider two neighboring time-dependent densities: $\rho_{t_{j-1}}$ and $\rho_{t_{j}}$. Similar to NCE, but in a dynamic setting, a time-dependent classifier $d_\theta$ is trained to distinguish samples from these two densities, where $\rho_{t_{j+1}}$ and $\rho_{t_j}$  play the role of the data and noise distribution, respectively. Therefore, like NCE, our approach transforms a density estimation problem into a binary classification problem. 
\end{rem}

\begin{rem}
    In some practical implementation $\Delta t_j$ can be chosen to be a constant, say $\Delta t$. The following discussion, including when we determine the stationary point for \Cref{eq:loss}, holds even if $\Delta t_j = \Delta t$.  
\end{rem}

\subsubsection{Stationary point of the proposed loss function}
We momentarily suppress the subscript $\thetaa$ and use $d^\ast$  to denote the minimizer of $\mathsf{J}$ in \Cref{eq:loss}. The following result identifies the stationary point $\mathsf{J}$: 
\begin{align}\label{eq:loss-stationary-point}
    &d^\ast \left(\x, \bar{t}_j , \Delta t_j \right) = \frac{\rho_{t_{j}}(\x)}{\rho_{t_{j-1}}(\x) + \rho_{t_{j}}(\x)} \qquad j \in \left\{ 1, 2, \ldots, N \right\}\nonumber \\
    \implies & \log \left( \frac{ \rho_{t_{j}}(\x)}{ \rho_{t_{j-1}}(\x)} \right) = \log \left(  \frac{d^\ast(\x, \bar{t}_j , \Delta t_j)}{1 - d^\ast(\x, \bar{t}_j , \Delta t_j)} \right) = f^\ast \left(\x, \bar{t}_j \right) \; \Delta t_j ,
\end{align}
where $f^\ast$ denotes the stationary point of the $\thetaa$ parameterized network corresponding to $d^\ast$. 

\begin{proof}[Derivation of \Cref{eq:loss-stationary-point}] 
We take variations of $d$ around the stationary point $d^\ast$ to determine the stationary point of $\mathsf{J}$, \ie we let 
\begin{equation}
    d(\x, t, \Delta t) = d^\ast(\x, t, \Delta t) + \epsilon g(\x, t, \Delta t) 
\end{equation} 
and substitute this in \Cref{eq:loss}. For this proof, we ignore the arguments of $d^\ast, g$ and $\rho_t$ to simplify notation. Also, we write $d(\x, \bar{t}_j, \Delta t_j), d^\ast(\x, \bar{t}_j, \Delta t_j)$ and $g(\x, \bar{t}_j, \Delta t_j)$ as $d_j, d^\ast_j$ and $g_j$, respectively.  Now, we start by substituting $d_j = d_j^\ast + \epsilon g_j$ in \Cref{eq:loss}, which leads to 
\begin{align}
    \mathsf{J} & = \sum_{j=1}^{N}  \left[ \int (d_j^\ast + \epsilon g_j)^2 \rho_{t_{j-1}}   \mathrm{d}\x  + \int \left( 1 - d_j^\ast -\epsilon g_j \right)^2 \rho_{t_{j}} \mathrm{d}\x  \right] \nonumber \\
    &= \sum_{j=1}^{N}  \left[ \int \left\{  {d_j^\ast}^2 + 2 \epsilon d^\ast g_j + \epsilon^2 g_j^2 \right\} \rho_{t_{j-1}}  \mathrm{d}\x  + \int \left\{ (1 - d_j^\ast)^2 - 2 \epsilon(1-d_j^\ast)g_j +  \epsilon^2 g_j^2 \right\} \rho_{t_{j}} \mathrm{d}\x  \right] \label{eq:stationary-deriv1}.
\end{align}
At the stationary point $d^\ast$
\begin{equation}
  \frac{ \mathrm{d}\mathsf{J}  }{ \mathrm{d}\mathsf{\epsilon} } \bigg\rvert_{\epsilon=0} = 0     \qquad \forall \text{ } g_j \text{ and } \forall j \in \{1, 2, \ldots, N\},
\end{equation}
Therefore, it follows that
\begin{equation}\label{eq:stationary-point-2}
	  \int 2 g_j    \left\{   d_j^\ast  \rho_{t_{j-1}}  -  (1-d_j^\ast) \rho_{t_{j}} \right\} \mathrm{d}\x  = 0  \qquad \forall \text{ } g_j \text{ and } \forall j \in \{1, 2, \ldots, N \},
\end{equation}
which means
\begin{align}
 d_j^\ast  \rho_{t_{j-1}}  -  (1-d_j^\ast) \rho_{t_{j}} = 0  & \implies  \frac{ \rho_{t_{j}}}{ \rho_{t_{j-1}}}  =   \frac{d_j^\ast}{1 - d_j^\ast}  \nonumber \\
&\implies \log \left( \frac{ \rho_{t_{j}}}{ \rho_{t_{j-1}}} \right) = \log \left(  \frac{d_j^\ast}{1 - d_j^\ast} \right) = f_j^\ast \Delta t_j, \label{eq:log-dens-difference}
\end{align}
where $f_j^\ast$ is $f^\ast \left(\x, \bar{t}_j\right)$. In deriving \Cref{eq:log-dens-difference}, we use \Cref{eq:sigmoid-reciprocal} to obtain the final quantity.
\end{proof}

\subsubsection{Estimating the time-dependent density \texorpdfstring{$\rho_t$}{}}\label{subsubsec:estimating-density}

Next, we show how to estimate $\rho_t$ using the time-dependent network $f^\ast$. We begin with \Cref{eq:loss-stationary-point}
\begin{equation}\label{eq:log-dens-evolution}
    \log \rho_{t}(\x) - \log \rho_{t-1}(\x) =  f^\ast(\x, \bar{t}_j) \Delta t_j.
\end{equation}
and sum both sides from $j=1$ to $N$, which yields
\begin{equation}\label{eq:telescopic-sum}
    \sum_{j=1}^{N} \Big[ \log \rho_{t}(\x) - \log \rho_{t-1}(\x) \Big] =  \sum_{j=1}^{N} f^\ast(\x, \bar{t}_j) \Delta t_j.
\end{equation}
The telescopic sum on the left hand side of \Cref{eq:telescopic-sum} further simplifies to yield
\begin{equation}\label{eq:midpoint-sum}
    \log \rho_T(\x) - \log \rho_0(\x) = \sum_{j=1}^{N} f^\ast \left(\x, \bar{t}_j \right) \Delta t_j .
\end{equation}
Similarly, for any time instance $t_{j^\prime}$, where $j^\prime \in \{2, \ldots, N\}$, the equation above can be modified to obtain $\log \rho_{t_j^\prime}$ as follows:
\begin{equation}\label{eq:midpoint-sum2} 
    \log \rho_{t_{j^\prime}}(\x) - \log\rho_{0}(\x) = \sum_{j=1}^{j^\prime} f^\ast (\x, t_j) \Delta t_j \qquad \forall \; j^\prime \in \{2, 3, \ldots, N\}.
\end{equation}
Moreover, for a value of $t \in (t_{j^\prime}, t_{j^\prime + 1})$, where no observations are available, we can estimate $\rho_t$ by adding
\begin{equation}\label{eq:midpoint-sum3}
    \log \rho_{t}(\x) - \log\rho_{t_j^\prime}(\x) = f^\ast\left(\x, \bar{t}_{j^\prime + 1}\right) (t - t_{j^\prime}) \quad \forall \; t \in (t_{j^\prime}, t_{j^\prime + 1}),
\end{equation}
to \Cref{eq:midpoint-sum2}. We obtain \Cref{eq:midpoint-sum3} by linearly interpolating between $\log \rho_{t_{j^\prime}}$ and $\log \rho_{t_{j^\prime+1}}$. As a result, the interpolation error is $\mathcal{O}(\Delta t_{j^\prime + 1}^2)$. 

In this work, we use \Cref{eq:midpoint-sum,eq:midpoint-sum2,eq:midpoint-sum3} to estimate the time-dependent density associated with a stochastic process, where the network $f^\ast$ is trained using observations of the stochastic process at different time instances.
We note that all the terms within the summation in \Cref{eq:midpoint-sum,eq:midpoint-sum2,eq:midpoint-sum3} are evaluated for a fixed value of $\x$, and can be evaluated in parallel on GPU processors, which adds to the overall computational efficiency. 

\begin{rem} \label{rem1:assumption}
The assumption that the density $\rho_0$ is analytically tractable can be relaxed. If the density $\rho_{t}$ is known at any point $t = t_{j^\dagger}$ along the stochastic process, the density at any other point along the process can be determined by shifting time by $t = t - t_{j^\dagger}$  and using \Cref{eq:midpoint-sum2} or \Cref{eq:midpoint-sum3}. Alternatively, if only samples are available from $\rho_0$, or at any point $t = t_{j^\dagger}$, then the proposed approach can estimate the probability density at the point. \Cref{sec:generative-model} discusses how the proposed approach can be used for time-independent or \emph{static} density estimation.

\end{rem}

\begin{rem}
We take expectations with respect to $\X \sim \rho_0$ on both sides of \Cref{eq:midpoint-sum2}. This will yield:
\begin{align}
    & \mathbb{E}_{\X \sim \rho_0} \left[  \log \hat{\rho}_{t_{j^\prime}} (\x) - \log \rho_{0} (\x)  \right] =  \mathbb{E}_{\X \sim \rho_0} \left[  \sum_{j=1}^{j^\prime} f^\ast\left(\x, \bar{t}_j \right) \Delta t_j   \right] \nonumber \\
    \implies & D_{\mathrm{KL}}\left( \rho_0 \mid\mid \rho_{t_{j^\prime}} \right) = - \sum_{j=1}^{j^\prime} \mathbb{E}_{\X \sim \rho_0} \left[ f^\ast\left(\x, \bar{t}_j \right) \right] \Delta t_j 
\end{align}
where $D_{\mathrm{KL}}\left( \rho_0 \mid\mid \rho_{t_{j^\prime}} \right)$ is the forward Kullback-Liebler (KL) divergence between $\rho_0$ and $\rho_{t_{j^\prime}}$. Similarly, the reverse KL divergence can be estimated by taking expectation with respect to $\X \sim \hat{\rho}_{t_{j^\prime}}$, \ie
\begin{equation}
    D_{\mathrm{KL}}\left( \rho_{t_{j^\prime}}  \mid\mid \rho_0 \right) = \sum_{j=1}^{j^\prime} \mathbb{E}_{\X \sim \hat{\rho}_{t_{j^\prime}}} \left[ f^\ast\left(\x, \bar{t}_j \right) \right] \Delta t_j. 
\end{equation}
Therefore, the proposed approach could be used to estimate the forward or reverse KL divergence between two distributions. This will be possible even when $\rho_0$ or $\rho_{t_{j^\prime}}$ can be sampled but cannot be evaluated. We don't explore this direction in this work, but we foresee that this will be helpful for, among other applications, the Bayesian design of experiments, in the context of inverse problems, to evaluate expected information gains. 
\end{rem}

\subsubsection{Interpretation of $f^\ast$}\label{subsubsec:interpretation}

Our formulation, including the loss \Cref{eq:loss} and the time-dependent architecture of the classifier lends itself to a natural connection between $f^\ast$ and $\partial \log \rho_t . \partial t$. To elucidate this connection we rewrite \Cref{eq:log-dens-evolution} as follows
\begin{equation}\label{eq:finite-difference-approximation}
    \frac{\log \rho_{t}(\x) - \log \rho_{t-1}(\x)}{\Delta t_j} =  f^\ast(\x, \bar{t}_j) .
\end{equation}
The left hand side is a $\mathcal{O}(\Delta t_j^2)$ approximation to $\partial \log \rho_{t}(\x)/ \partial t$. Therefore, we may write 
\begin{equation}\label{eq:second-order-approx-f}
    \frac{\partial \log \rho_{t}(\x)}{\partial t} \Bigg\rvert_{t= \bar{t}_j} = f^\ast(\x, \bar{t}_j)  + \mathcal{O}(\Delta t_j^2) \quad \forall \, j \in \{1,2, \ldots, N\}
\end{equation}
and recognize its implication that, for a finite value of $\Delta t_j$, $f^\ast$ provides a second-order approximation to the partial time derivative of $\rho_{t}$ when both quantities are evaluated at $t = \bar{t}_j \,\; \forall \, j \in \{1, 2, \ldots, N\}$.

\section{Density estimation and sampling}\label{sec:generative-model}
The previous section demonstrates how a time-dependent classifier can learn the time-dependent probability density of a multivariate stochastic process from sample path data. A related, and perhaps more common, task in statistical computing involves estimating the unknown probability density $\rho_{\mathrm{data}}$ of a $n$-dimensional random variable $\X$ from a sample of $N_\mathrm{data}$ iid realizations $\{ \x^{(1)}, \x^{(2)}, \ldots, \x^{(N_{\mathrm{data}})}\}$~\cite{soize2016data}, which we denote using $\mathcal{D}$. This task is known as density estimation, and itself forms the basis of other tasks like computing population parameters, generating new samples, and detecting rare events. Density estimation and sampling are also fundamental to various applications like uncertainty quantification and generative modeling. This section extends the proposed approach to tackle time-independent or static density estimation and sampling.

\subsection{Stochastic interpolants} 

In static density estimation, the time-independent density $\rho_{\mathrm{data}}$ is of interest. We start by constructing a stochastic process with a time-dependent density for which the terminal condition $\rho_{T}$ is the target density $\rho_{\mathrm{data}}$. Without a loss in generality, we set $T= 1$. 

\begin{defn} 
Let $\X_0 \sim \rho_0$ and $\X_{1} = \X \sim \rho_{\mathrm{data}} = \rho_1$. A stochastic interpolant between $\X_0$ and $\X_1$ is a stochastic process satisfying
\begin{equation}\label{eq:stochastic-interpolant-1}
	\X_t = \mathcal{I}_t(\X_0, \X_1) \qquad\qquad \forall \;\; t \in [0, 1],
\end{equation}
where $\mathcal{I}$ satisfies the boundary conditions $ \mathcal{I}_{t = 0}(\X_0, \X_1) = \X_0$ and $ \mathcal{I}_{t=1}(\X_0, \X_1) = \X_1$~\cite{albergo2023stochastic,albergo2022building}. 
\end{defn}
\noindent There are many ways to construct $\mathcal{I}$; see Table 4 in \cite{albergo2023stochastic}. In this work, we adopt the \emph{linear interpolant}~\cite{albergo2022building}
\begin{equation}\label{eq:linear-interpolant}
	\X_t = (1 - t) \X_0 + t \X_1 \qquad\qquad \forall \;\; t \in [0, 1]
\end{equation}
It is easy to verify that \Cref{eq:linear-interpolant} satisfies the property of a stochastic interpolant, ensuring any arbitrary density $\rho_0$ evolves to the desired density $\rho_1$ in finite time. Therefore, we choose $\rho_0$ to be a tractable density, like the standard normal distribution in $\Re^n$, which we can evaluate and efficiently sample. Herein, we refer to $\X_0$ as the latent variable and $\rho_0$ as the latent distribution. 

The linear interpolant constructs a continuous path between realizations drawn from the latent distribution $\rho_0$ and the target data distribution $\rho_1$. We choose the linear interpolant because it is simple to implement and sufficient for density estimation. The classifier learns how the log-density evolves along the paths induced by the interpolant, which need not be physically meaningful. In general, the estimated density is not biased by the specific choice of interpolant. However, in more complex settings, for example, when data concentrates on a low-dimensional manifold, the method may benefit from carefully designed interpolants that remain constrained on the manifold~\cite{kapusniak2024metric}. We leave the exploration of more sophisticated interpolants to future work.

\subsection{Density estimation}

To train the time-dependent classifier, we discretize the time interval $[0,1]$, which entails choosing the time instants $t_1$ through $t_T$ such that $t_0 = 0 <  t_1 < t_2 < \cdots < t_{N} = 1$. These may be chosen on a linear scale such that $\Delta t_j = 1/N \;\; \forall j$, or on a logarithmic scale such that $t_{j+1}/t_j = \gamma \;\; \forall j \geq 1$ and $\gamma > 0$. 
Next, the objective $\mathsf{J}$ is approximated using \Cref{eq:loss-MCS} but with $\x^{(i)}_{t_j} $ now formed during training using \Cref{eq:linear-interpolant} as follows,
\begin{equation}\label{eq:linear-interpolant-2}
	\x^{(i)}_{t_j} = (1 - t_j) \x_{0}^{(i)} +  t_j \x^{(i)} ,
\end{equation}
where the latent variable $\x_0$ is sampled randomly from $\rho_0$ every time $\x_{t_j} $ must be formed, while $\x$ is sampled uniformly from $\mathcal{D}$. Therefore, \Cref{eq:linear-interpolant-2} results in an unpaired dataset. Once the classifier has been trained, we estimate $\rho_{\mathrm{data}}$ using \Cref{eq:midpoint-sum} as follows
\begin{equation}\label{eq:density-estimate}
	\log \rho_{\mathrm{data}}(\x) = \log \rho_1(\x) = \log \rho_{0}(\x) + \sum_{j=1}^{N} f^\ast \left(\x, \bar{t}_j\right) \Delta t_{j} 
\end{equation}
where $f^\ast$ is the optimally trained network within the classifier, $\bar{t}_j = t_{j-1} + \Delta t_j/2$ and $\Delta t_j = t_{j} - t_{j-1}$. 

\subsection{Sampling}

To generate new samples from $\rho_{\mathrm{data}}$, one can use the expression obtained by exponentiating \Cref{eq:density-estimate} in any Markov chain Monte Carlo (MCMC) simulation method. However, gradient-based MCMC methods are more suitable for exploring high-dimensional spaces. Remarkably, the score function, defined as $\nabla \log \rho_{\mathrm{data}}$~\cite{dasgupta2025conditional}, can be computed using
\begin{equation}\label{eq:score-function}
	\nabla_{\x} \log \rho_{\mathrm{data}}(\x) = \nabla_{\x} \log \rho_{0}(\x) + \sum_{j=1}^{N} \nabla_{\x} f^\ast \left(\x, \bar{t}_j\right) \Delta t_{j} , 
\end{equation}
where $\nabla f^\ast$ represents the gradient of the output of the network $f^\ast$ with respect to its input $\x$. Automatic differentiation~\cite{baydin2018automatic} enables the efficient computation of $\nabla_{\x} f^\ast$. So, \Cref{eq:score-function} can be used with gradient-based MCMC methods such as the unadjusted Langevin Monte Carlo algorithm (ULA) or the Hamiltonian Monte Carlo (HMC) algorithm and their variants to sample high-dimensional spaces; see \cite{girolami2011riemann} for a concise introduction to both algorithms. 

While our method enables efficient computation of the score function $\nabla_{\x} \log \rho_{\mathrm{data}}(\x)$ via automatic differentiation, there are practical considerations when using it within gradient-based MCMC methods. 
We observe empirically that careful tuning of sampling hyperparameters improves mixing and acceptance rates in high dimensions. Mainly, we observe that reducing the step size and increasing the number of time steps $N$ helps improve overall sample quality. Additionally, initializing Markov chains near regions of high data density by perturbing training data points ensures individual chains reach their steady state quickly. This is a prevalent practice commonly known as \emph{data initialization} in the literature~\cite{soize2016data,gottwald2024stable}.

\subsection{Comparison with other generative models}
Any mechanism that can generate new realizations from an unknown data-generating distribution with density $\rho_{\mathrm{data}}$, using only realizations $\{ \x^{(1)}, \x^{(2)}, \ldots, \x^{(N_{\mathrm{data}})}\}$ for learning, is a generative model. Therefore, the proposed approach acts as a generative model when we use it to approximate $\rho_{\mathrm{data}}$ and then sample it. 
We devote this discussion to contrasting the proposed approach against popular generative models focusing on what each type of generative model tries to approximate. 
\begin{itemize}[leftmargin=*]
    \item Generative adversarial networks~\cite{goodfellow2020generative} (GANs) typically push forward a low-dimensional latent vector to a realization from the target distribution. This requires a generator and discriminator pair, trained using an adversarial objective typically susceptible to training instabilities. We did not observe such instabilities in our experiments. Moreover, GANs cannot be used to evaluate $\rho_{\mathrm{data}}$. 
    \item Normalizing flows~\cite{kobyzev2020normalizing} use invertible neural networks to construct a push-forward map that transports realizations from the latent distribution to corresponding realizations from the target distribution. However, as we already alluded to in the \Cref{sec:introduction}, invertible neural networks are specialized architectures that have significant memory footprints that can be overwhelming for high-dimensional applications. In contrast, the network $f$ can be flexibly constructed because it is a vector-to-scalar map that is not invertible. This also opens up opportunities for working with pre-trained embeddings when available. For instance,  it is possible to construct $f_{\thetaa} = f_{1, \thetaa} \circ f_2 $, where $f_2$ can be a pre-trained feature extractor and $f_{1, \thetaa}$ is a trainable classification head. We envisage that this flexibility will be really important in high-dimensional applications, such as when working with structured data such as spatial fields or natural images. Indeed, we exploit this flexibility in \Cref{subsec:outlier-detection-results}. 
    
    \item Flow-based generative models~\cite{lipman2022flow,lipman2024flow,albergo2022building,albergo2023stochastic}, including diffusion models~\cite{song2020score,dasgupta2025unifying}, learn a velocity field with which realizations from a latent distribution may be evolved to obtain realizations from the target distribution. In this case, the generator is the probability flow-ODE and the total derivative of the log-density satisfies the continuity equation with the same velocity field. Different flow-based generative models vary in their choice of the latent distribution, and the stochastic process (analogous to \Cref{eq:stochastic-interpolant-1}) that describes how realizations from the latent distribution and the data distribution are related.  For example, diffusion models use linear SDEs with suitable drift and diffusion terms such that data distribution is progressively convolved by a Gaussian kernel of increasing variance~\cite{song2020score}. In contrast, $f$ approximates the partial derivative of the log-density with respect to time of the stochastic process induced by the stochastic interpolant. Although these methods permit log-density computations, evaluating the log-density requires expensive integration of the continuity equation. In contrast, we evaluate the log-density by evaluating $f$ at various points along $t$, while $\x$ is held constant in \Cref{eq:density-estimate}.
    
    \item Energy-based models~\cite{song2021train} (EBMs) explicitly approximate the log-density using a neural network, \ie $\rho_{\mathrm{data}} \propto \exp(g_{\thetaa}(\x))$. NCE remains a popular framework for training EBMs~\cite{song2021train}. However, choosing the contrastive term is non-trivial, and the resulting performance of the EBM is sensitive to that choice. Recent advances in contrastive estimation use the energy-based model from the previous iteration to generate realizations to approximate the second term~\cite{song2021train}. This requires MCMC simulations~\cite{du2020improved} or flow-based modeling~\cite{gao2020flow} to generate samples from the energy-based model during training, which can be computationally inefficient. In contrast, \Cref{eq:linear-interpolant,eq:linear-interpolant-2} can be evaluated efficiently during training.  
\end{itemize}
We compare our approach, in terms of the quality of the generated samples, with diffusion models and find that our approach is competitive.  However, a rigorous comparison against other generative models on density estimation tasks and sample quality is beyond the scope of the current work; we leave that for a future study. 

\section{Numerical examples}\label{sec:numerical_examples}

\subsection{Results on estimating the time-dependent density of stochastic processes using the proposed approach}  \label{subsec:results-javier}

We demonstrate the ability of the time-dependent classifier to estimate the time-dependent density of stochastic processes arising from dynamical systems driven by white noise excitation. These examples, adapted from \cite{tabandeh2022numerical}, consist of synthetic stochastic processes where the true time-evolving densities are not known in closed form. Specifically, these examples feature a duffing and hysteretic oscillator subject to white noise excitation, leading to two- and three-dimensional systems. We provide additional details regarding the architecture of $f$ and its training in \Cref{appsubsec:time-dependent-experiments}.  

\subsubsection{Duffing oscillator}

The first representative case study, the duffing oscillator, is a nonlinear dynamical system frequently encountered in various fields such as plasma physics, electronic systems, and communication theory \cite{ueda1979randomly,hammad2020new}. The system dynamics are governed by the following second-order stochastic differential equation
\begin{equation}
\ddot{y}_t + 2\zeta \omega_0 \dot{y}_t + \omega_0^2 \left( y_t + \epsilon_{\text{nl}} y_t^3 \right) = \xi_t,
\label{eq:duffing}
\end{equation}
where $\zeta$ denotes the damping ratio, $\omega_0$ is the undamped natural frequency, $\epsilon_{\text{nl}}$ quantifies the degree of nonlinearity, and $\xi_t$ is a Gaussian white noise process with power spectral density $s_0$. We define the state vector as $\x_t\transpose = [ y_t, \dot{y}_t ]\transpose$, \ie $n=2$, and consider the parameter values $\zeta = 0.25$, $\omega_0 = 1$, $\epsilon_{\text{nl}} = 1$, and $s_0 = 0.5$. The initial condition $\rho_{0}$ is assumed to be a bivariate Gaussian probability density function with zero mean, unit variance for both state-space coordinates, and a correlation coefficient of 0.5 between them. We obtain the reference solution for the transient behavior by estimating the time-dependent density $\rho_t(\x)$ where necessary, using kernel density estimation (KDE) applied to trajectories generated using Monte Carlo simulations. To realize a trajectory, we numerically integrate \Cref{eq:duffing} using a Runge--Kutta method with a fixed time step of 0.005 and obtain the state space for the time interval $[0, 0.8]$ corresponding to a realization of the noise process $\xi_t$. The goal is to estimate the density $\rho_t \; \forall \, t \in [0, 0.8]$ of the 2-dimensional state vector. 
In this example, the dataset is paired since every trajectory is observed at every time instant $t_j = j\Delta t \;\; \forall j \in \{1, 2,  \ldots, 120\}$ where $\Delta t = 0.1$, $N = 160$ and $t_N  = 0.8$. Using the available data, we train the classifier and subsequently use the trained classifier to estimate $\rho_t$. 
\begin{figure}[!b]
    \centering
    \includegraphics[width=6in]{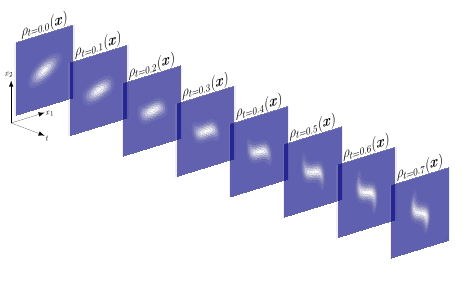}
    \caption{Snapshots of the time-dependent density $\rho_t$ for the Duffing oscillator estimated using the proposed approach}
    \label{fig:duffing-osc-evo}
\end{figure}

\begin{figure}[t]
    \centering
    \hspace{-0.5cm}
    \includegraphics[width=1.0\linewidth]{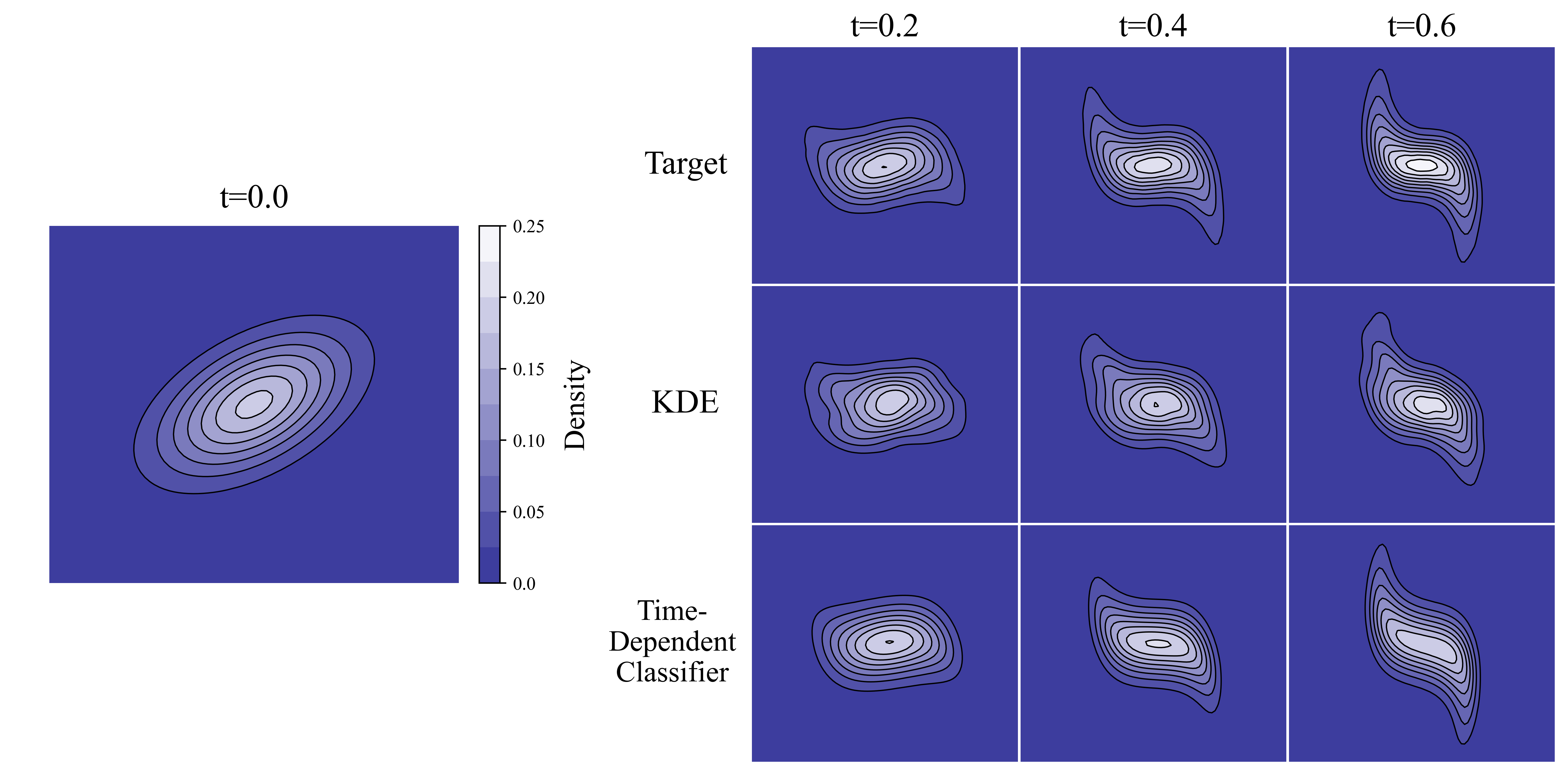}
    \caption{Time-dependent density for the Duffing oscillator. Left: initial density. Right: Density estimates at $t= \{0.2, 0.4, 0.6\}$. The top row shows the target density, the middle row shows the density obtained using kernel density estimation with 1,000 points and Silverman's bandwidth~\cite{silverman2018density}, and the bottom row shows the density estimated using the proposed approach with 1,000 samples per time step.}
    \label{fig:duffing_osc}
\end{figure}

\Cref{fig:duffing-osc-evo} illustrates the time evolution of $\rho_t$ for the Duffing oscillator whereas \Cref{fig:duffing_osc} shows a select few snapshots. In  \Cref{fig:duffing_osc}, the figure on the left shows the initial density and the panels on the right show the estimated densities at $t= \{0.2,0.4,0.6\}$. The top row presents reference density obtained using kernel density estimation with 500,000 samples. The middle row displays kernel density estimates computed from 1,000 simulated trajectories using Silverman’s bandwidth selection rule~\cite{silverman2018density}. The bottom row shows the densities inferred by the proposed time-dependent classifier. We observe that the time-dependent classifier closely matches the target distributions over time.

\subsubsection{Hysteretic oscillator}

We investigate a nonlinear oscillator that exhibits hysteretic dynamics, a behavior that is commonly observed in many domains including biomechanics, optics, electronic systems, ferroelectric materials, magnetic systems, and structural mechanics \cite{ismail2009hysteresis}. In structural systems, hysteretic materials provide restoring forces while dissipating energy under external loads. The following second-order stochastic differential equation describes the evolution of a hysteretic oscillator,
\begin{equation}
\ddot{y}_t + 2\zeta \omega_0 \dot{y}_t + \omega_0^2 \left[ \alpha_e y_t + (1 - \alpha_e) z_t \right] = \xi_t,
\label{eq:hyst_sde}
\end{equation}
where \( \alpha_e \in [0,1] \) represents the ratio between post- and pre-yield stiffnesses, $\omega_0$ is the natural frequency of the system, \( z_t \) is the hysteretic contribution to the restoring force, and \( \xi_t \) is Gaussian white noise with power spectral density \( s_0 \). The hysteretic term \( z_t \) evolves according to the Bouc--Wen hysteresis model \cite{wen1976method,wen1980equivalent}, which introduces a nonlinear differential equation of the form:
\begin{equation}
\dot{z}_t = A \dot{y}_t -\gamma |\dot{y}_t| |z_t|^{\nu-1} z_t - \beta \dot{y}_t |z_t|^{\nu} ,
\end{equation}
where \( \gamma, \nu, \beta, A \) are model parameters that shape the force--displacement hysteresis loop and govern its smoothness.

To recast this system into first-order state space form suitable for numerical integration, we define the augmented state vector \( \x_t\transpose = [ y_t, z_t, \dot{y}_t]\transpose\). Therefore, $n=3$ in this example.  
Further, we consider the parameters values \( \zeta = 0.05 \), \( \omega_0 = 1 \), \( \alpha_e = 0.01 \), \( \gamma = 1 \), \( \beta = 1 \), \( \nu = 1 \), \( A = 1 \), and \( s_0 = 0.5 \). The initial distribution is Gaussian with zero mean, unit variance in each component, and a pairwise correlation coefficient of 0.8. The goal is to estimate the time-evolving density $\rho_t(\x)$, which characterizes the transient response. We acquire the data necessary for training the classifier by numerically integrating \Cref{eq:hyst_sde} using a Runge--Kutta method with a fixed time step of 0.005 in the time interval $[0, 0.8]$. As a result, $N = 160$ and $t_N = 0.8$ for this example. Using this data, we train the time-dependent classifier, with a constant $\Delta t =  0.1$, to estimate the partial derivative of $\log \rho_t$.
\begin{figure}[!b]
    \centering
    \hspace{-0.5cm}
    \includegraphics[width=1.0\linewidth]{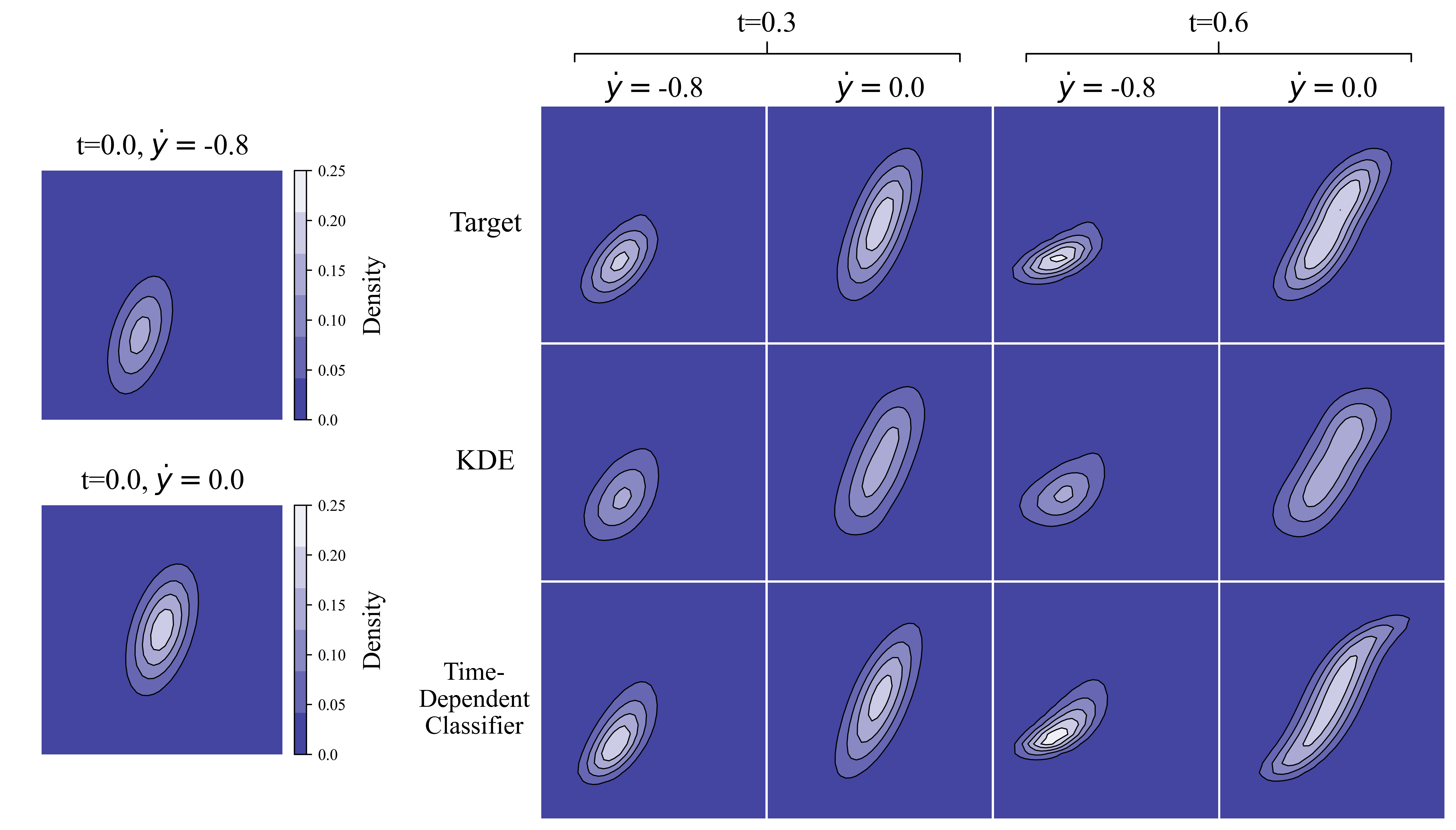}
    \caption{Time-dependent density for the hysteretic oscillator. Left: initial density. Right: Density estimates in the $y-z$ plane for $\dot{y}=-0.8$ and $\dot{y}=0.0$, at $t= \{0.2, 0.3\}$. The top row shows the target density, the middle row shows the density obtained using kernel density estimation with 10,000 points  and Silverman's bandwidth~\cite{silverman2018density}, and the bottom row shows the density estimated using the proposed approach with 10,000 samples per time step}
    \label{fig:hysteretic_osc}
\end{figure}

\Cref{fig:hysteretic_osc} illustrates snapshots of the probability density $\rho_t$. The figure on the left shows the initial density for two separate values of $\dot{y}$. The panels on the right compare the estimated joint density between $y$ and $z$ at two representative time steps $t= \{0.3,0.6\}$ for two separate values of $\dot{y}$. The top row presents the reference density obtained using kernel density estimation with 500,000 samples. The middle row displays kernel density estimates computed from 10,000 samples using Silverman’s bandwidth selection rule~\cite{silverman2018density}. The bottom row shows the densities inferred by the proposed time-dependent classifier from the same number of observations at all time instances. 
It is observed that qualitatively, the time-dependent classifier matches the target distribution more closely than the kernel density estimate obtained using the same number of samples. 

\subsubsection{Discussion}

The proposed approach captures the transient behavior of the probability densities associated with the Duffing and hysteretic oscillator examples, using sample path data only, and, more significantly, without any knowledge of the underlying physics, such as the drift and diffusion terms associated with the underlying SDE. Further, \Cref{tab:l2-error-dynamical-systems} compares the $L_2$ error between the densities estimated using the proposed approach and KDE, with the same number of samples in the training dataset. Note that we perform the KDE independently at different instants of $t$. In contrast, our approach estimates the time-dependent density $\rho_t$ using \Cref{eq:midpoint-sum2}. For the 2-dimensional problem, the $\mathcal{L}_2$ error corresponding to our approach is marginally larger than KDE; it is well known that KDE performs well in low-dimensional settings. However, as the dimension of the problem increases, such as in the 3-dimensional example involving the hysteretic oscillator, the KDE approach does not scale, and the classifier-based approach outperforms KDE. Further, as we show in the following section, for cases where the measure concentrates around a low-dimensional manifold, the KDE approach yields inaccurate results. Overall, \Cref{subsec:results-javier} shows that the proposed approach is a promising method for modeling time-dependent probability densities associated with stochastic processes, like those arising from dynamical systems driven by stochastic excitations.  
\begin{table}[H]
\centering
\caption{$L_2$-norm of the difference between the target density $\rho_t(\boldsymbol{x})$ and the density estimated using KDE (with Silverman's rule) and the time-dependent classifier. The target density is computed using KDE with 500,000 samples}
\label{tab:l2-error-dynamical-systems}
\begin{tabular}{l c c c @{\extracolsep{5pt}}cc c c c}
\toprule[1.5pt]
\multirow{3}{*}{Method} & \multicolumn{3}{c}{Duffing Oscillator} & \multicolumn{3}{c}{Hysteretic Oscillator} \\
\cline{2-4} \cline{5-8}
& \multirow{2}{*}{$t=0.2$} &\multirow{2}{*}{$t=0.4$}&\multirow{2}{*}{$t=0.6$}& \multicolumn{2}{c}{$t=0.3$}  & \multicolumn{2}{c}{$t=0.6$}  \\
\cline{5-6} \cline{7-8}
& & & & $\dot{y}=-0.8$  & $\dot{y}=0.0$ & $\dot{y}=-0.8$ & $\dot{y}=0.0$ \\
\midrule[1pt]
KDE & 0.036 & 0.045 & 0.051 & 0.017 & 0.021 & 0.030 & 0.030 \\
Classifier & 0.036 & 0.047 & 0.063 & 0.017 & 0.008 & 0.020 & 0.026 \\
\bottomrule[1.5pt]
\end{tabular}
\end{table}

\subsection{Results using the proposed approach for density estimation and sampling}\label{sec:results-dens-est-sampling}

To evaluate the efficacy of the proposed approach for density estimation and sampling, we present a series of experiments across datasets of increasing  dimension. We choose these examples to test the method's ability to capture complex structures underlying the data generation process in moderately high-dimensional spaces. Specifically, we consider synthetic 2D datasets (concentric circles, two moons, and checkerboard patterns) with disconnected supports as well as higher-dimensional examples where the data concentrates on lower-dimensional manifolds (compared to the ambient dimensionality of the problem) or exhibits strong correlations between variables. For each case, we estimate the data density from finite samples, use gradient-based MCMC to generate new samples from the learned model, and compare the generated samples to the original data. We relegate details regarding the architecture of $f$ and its training to \Cref{appsubsec:models}. Unless we mention otherwise, we use $N = $ 10 steps for the two-dimensional examples and $N=$ 20 steps for the other examples, with successive time steps chosen uniformly on a logarithmic scale. We also compare the sample quality with diffusion models~\cite{song2020score,dasgupta2025unifying}, widely considered to be the state-of-the-art in generative modeling. We provide additional details regarding the diffusion models used for these examples in \Cref{appsubsec:models}. 

\subsubsection{Concentric circles} 

\begin{figure}[!b]
	\centering
	\includegraphics[height=1.8in]{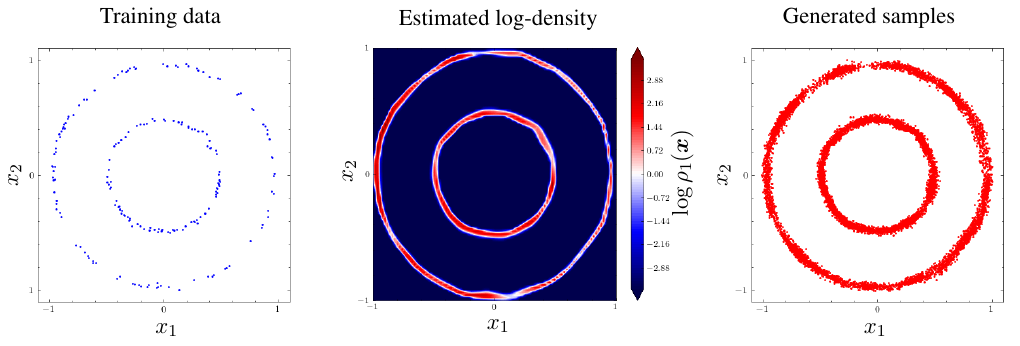}
	\caption{Training data (left), estimated density (middle), and generated samples (right) for the concentric circles dataset}
	\label{fig:circles-training-samples}
\end{figure}

This dataset consists of  $N_{\mathrm{data}} = 200$ realizations of a 2-dimensional random vector $\X$ sampled from the following data generating distribution
\begin{equation}\label{eq:circles}
	\X = (1 - p/2) \left[ \cos \theta , \sin \theta \right]\transpose + 0.01\Z \quad,
\end{equation}
where $\theta \sim \mathcal{U}(0, 2\pi)$ is a uniformly distributed random variable between 0 and $2\pi$, $\Z \sim \mathcal{N}(0, \mathbb{I}_2)$ is a bivariate standard normal variable and $p \sim \text{Bern}(0.5)$ is a Bernoulli random variable that is 0 or 1 with a probability of 0.5. The first column in \Cref{fig:circles-training-samples} shows the training data for this problem. 
\begin{figure}[!b]
	\centering
	\includegraphics[height=1.6in]{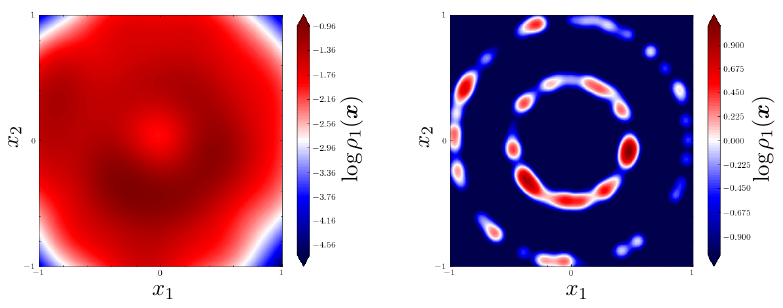}
	\caption{Log-density estimates for the concentric circles dataset obtained using kernel density estimation with Silverman's bandwidth~\cite{silverman2018density} (left) and bandwidth equal to 0.09 (right)}
	\label{fig:circles-kde-estimates}
\end{figure}

The second column in \Cref{fig:circles-training-samples} shows the estimated log-density $\log \rho_{1}$ obtained using \Cref{eq:density-estimate} and $N = 10$ time steps chosen uniformly on a logarithmic scale. In comparison, \Cref{fig:circles-kde-estimates} shows log-density values corresponding to kernel density estimates for $\rho_{\mathrm{data}}$ obtained using squared exponential kernels with two different bandwidths. Although $\X_1$ is infinitely supported, kernel density estimation struggles because the empirical distribution largely concentrates over a thin band around the two concentric circles. The proposed approach gives a density $\rho_1$ that closely matches the data density $\rho_{\mathrm{data}}$, in part due to the strong nonlinear approximation ability of the network $f$,  which is also evident in the generated samples. The third column in \Cref{fig:circles-training-samples} shows 10,000 realizations generated using ULA with a step size of 0.0001 and 200 steps. We seed individual Markov chains with realizations of a random variable uniformly distributed over $[-1, 1] \times [-1, 1]$. Note how new samples are generated even in locations where there are gaps in the outer ring where training data is absent.

\begin{figure}[t]
    \centering
    \begin{tikzpicture}
        \node[yshift=10pt] (A) at (0,0) {\includegraphics[height=2.2in,trim={30 0 0 0},clip]{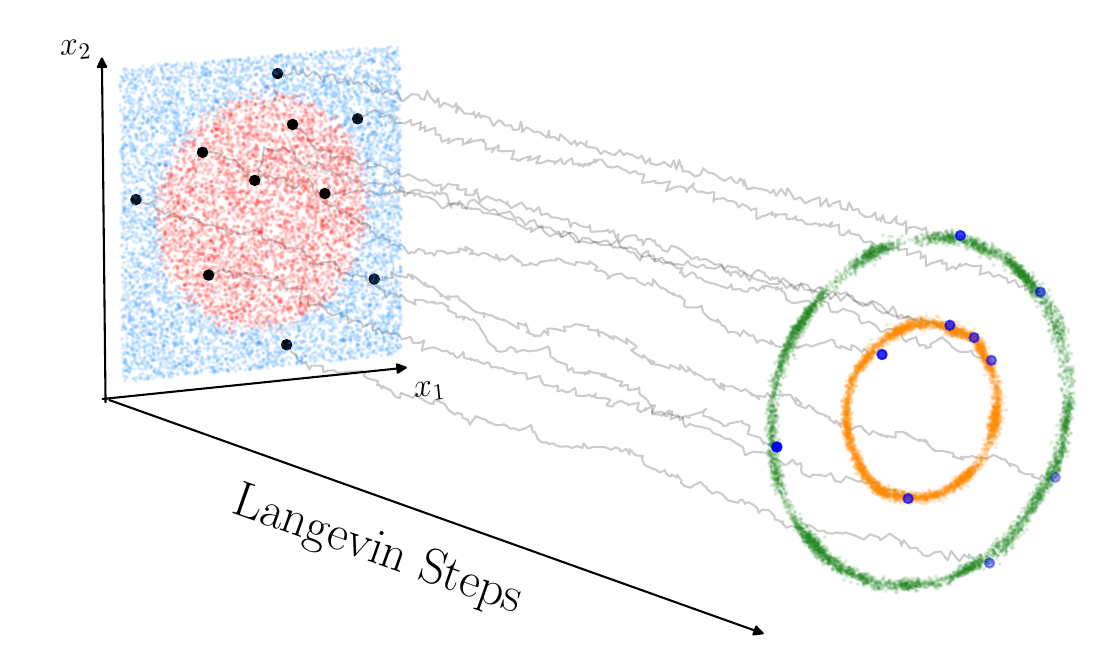}};
        \node[] (B) at (8,0) {\includegraphics[height=2.4in]{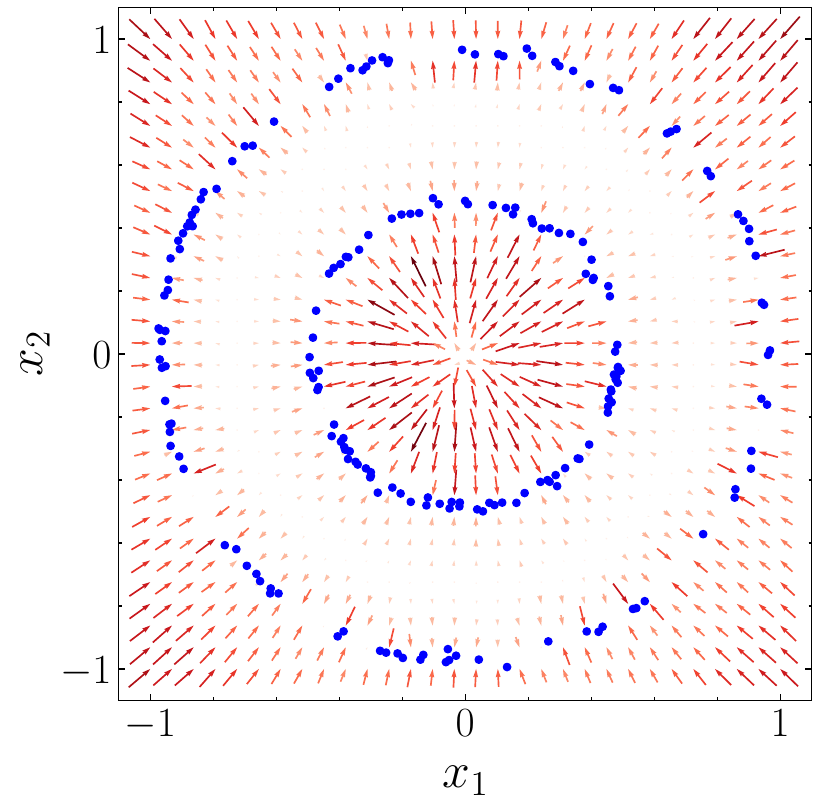}};
        \node[] at (0, -3.5) {(a)};
        \node[xshift=10pt] at (8, -3.5) {(b)};
    \end{tikzpicture}
    \caption{(a) Realizations of a random variable uniformly distributed over $[-1, 1] \times [-1, 1]$ ($\bullet$) that seed independent Markov chains (gray lines) are evolved into independent realizations of $\X$ (\textcolor{blue!60}{$\bullet$}). (b) Score function $\nabla_{\x} \log \rho_1(\x)$ for the concentric circles dataset, where deeper colors indicate larger magnitudes of the score}
	\label{fig:circles-ULA-steps}
\end{figure}

\Cref{fig:circles-ULA-steps}(a) shows how new realizations evolve through successive steps in five independent ULA chains. Even initial points in extremely low-density regions of $\rho_1$ evolve to realizations that concentrate on the two concentric circles corresponding to the two modes of the data distribution. Initial points around the periphery (blue points) evolve to points on the outer circle, whereas points close to the center (red points) move to the inner circle. Thus, the score function, derived from the density estimated using the proposed approach and shown in \Cref{fig:circles-ULA-steps}(b), is useful in sparse regions of the training data. 
\begin{table}[!b]
	\centering
	\caption{Regularized OT distance between the generated samples and the test data for the concentric circles data}
	\label{tab:regOTdist-circle}
	\begin{tabular}{lcccc}
	\toprule[1.5pt]
	Method & \makecell{Proposed\\approach} & \makecell{KDE\\(Silverman's)} & \makecell{KDE\\(Bandwidth=0.09)} & \makecell{Diffusion\\model} \\
        \midrule[1pt]
        OT distance & 0.065 & 0.168 & 0.104 & 0.076 \\
	\bottomrule[1.5pt]
	\end{tabular}
\end{table}

\begin{figure}[t]
	\centering
	\includegraphics[height=1.8in]{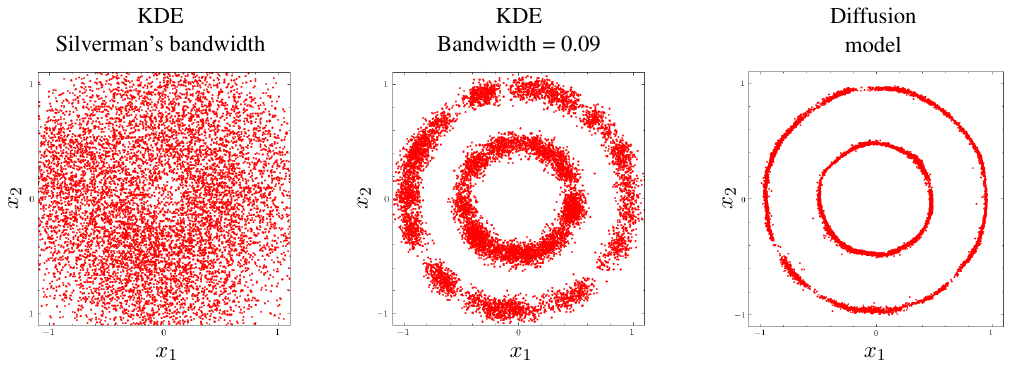}
	\caption{Samples generated from the density estimated using KDE --- Silverman's bandwidth~\cite{silverman2018density} (left) and bandwidth equal to 0.09 (middle) --- and using diffusion models (right)}
    \label{fig:circles-kde-diffusion-samples}
\end{figure}

\Cref{tab:regOTdist-circle} reports the regularized optimal transport (OT) distance between the generated samples and a test set (realizations not used for training) using the Sinkhorn-Knopp algorithm~\cite{cuturi2013sinkhorn} with the regularization parameter set to 0.01. We also report the regularized OT distance between samples generated from the KDE estimate for $\rho_{\mathrm{data}}$ and a trained diffusion model, and the test set for comparison. Smaller values mean that samples from the estimated density and the true density of the data-generating distribution are closer. \Cref{tab:regOTdist-circle} reflects \Cref{fig:circles-training-samples,fig:circles-kde-diffusion-samples}, where \Cref{fig:circles-kde-diffusion-samples} shows the samples generated using diffusion models and the density estimated via KDE, and quantifies that the proposed approach is better than KDE and competitive with diffusion models. Herein, we will only compare the sample quality with diffusion models because KDE does not yield a good estimate for the log-density in the examples we consider. 

Next, we study the effect of varying $N$ on the sample quality and sampling time. \Cref{tab:regOTdist-circle-varyNT} reports the regularized OT distance between the generated realizations and a test set, and the time required to sample 10,000 realizations as $N$ increases, averaged over 10 independent runs. We report the wall times necessary on an NVIDIA RTX 4090 GPU. Moreover, we discretize $[0, 1]$ uniformly into $N$ chunks for these experiments. \Cref{tab:regOTdist-circle-varyNT} shows that the quality of the generated samples improves as $N$ increases. However, with a larger $N$ the sampling time also increases, from 1 s for $N=1$ to 42 s for $N=100$. However, the training time for a fixed number of epochs and batch size is the same, around 1.5 minutes, irrespective of the value of $N$. Comparing \Cref{tab:regOTdist-circle,tab:regOTdist-circle-varyNT} for $N=10$, we observe that using time steps distributed uniformly on a logarithmic scale leads to better performance than using linearly spaced steps. This improvement arises because the logarithmic schedule allocates smaller time steps near $t=1$, where the evolving density $\rho_t$ approaches the target data distribution $\rho_{\mathrm{data}}$. Since the distribution becomes more complex near the terminal time, taking finer steps in this region is beneficial. Therefore, we choose to work with time steps chosen uniformly on a logarithmic scale for the remaining experiments. 
\begin{table}[H]
\centering
\caption{Regularized OT distance between the generated samples and the test data, and the sampling time for varying values of $N$, averaged across 10 independent runs, for the concentric circles data}
\label{tab:regOTdist-circle-varyNT}
\begin{tabular}{lcccccccccccc}
\toprule[1.5pt]
    $N$ & 1& 2& 5& 10& 20& 30& 40& 50& 100\\
    \midrule[1pt]
    OT distance & 0.139 & 0.122 &  0.079 & 0.081 & 0.080 & 0.073 & 0.072 & 0.064 & 0.065\\
    Sampling times & 1s  & 2s  & 3s & 5s & 9s & 13s & 18s & 21s & 42s\\
    \bottomrule[1.5pt]
\end{tabular}
\end{table}

\subsubsection{Two moons}

\begin{figure}[t]
	\centering
	\includegraphics[height=1.8in]{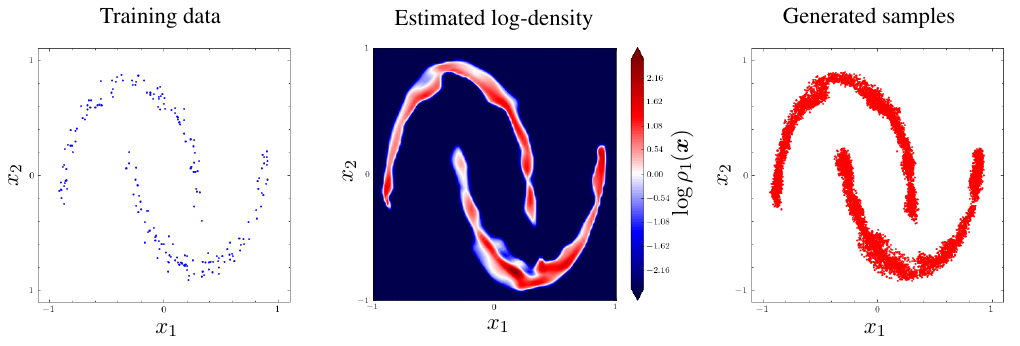}
	\caption{Training data (left), estimated density (middle) using the proposed approach, and generated samples (right) for the two moons dataset}
	\label{fig:moons-training-samples}
\end{figure}

This dataset also consists of $N_{\mathrm{data}} = 200$ realizations of a 2-dimensional random variable sampled from the following data-generating distribution
\begin{equation}\label{eq:two-moons}
	\X = 2 \left[ (\cos \theta)^{1-p}  (1 - \cos \theta)^{p} , (\sin \theta)^{1-p}  (0.5 - \sin \theta)^{p} \right]\transpose + 0.1\Z - 1,
\end{equation}
where $\theta \sim \mathcal{U}(0, \pi)$, $\Z \sim \mathcal{N}(0, \mathbb{I}_2)$ and $p \sim \text{Bern}(0.5)$. The first column in \Cref{fig:moons-training-samples} shows the training data for this problem. The second column in \Cref{fig:moons-training-samples} shows the estimated log-density $\log \rho_{1}$ obtained using \Cref{eq:density-estimate} and $N = 10$ time steps chosen uniformly on a logarithmic scale. The proposed method successfully recovers the crescent-shaped structures of both modes, preserving their sharp boundaries and maintaining the separation between them. However, the boundaries are not entirely smooth, reflecting the lack of training data in those regions. For example, a noticeable kink appears along the convex edge near the zenith of the upper moon, corresponding to an abrupt sparsity in the corresponding region of the training data. The third column in \Cref{fig:moons-training-samples} also shows 10,000 realizations generated using ULA with a step size of $0.0001$ and 200 steps. Like the previous experiment, we seed individual Markov chains with realizations of a random variable uniformly distributed over $[-1, 1] \times [-1, 1]$. The generated realizations closely follow the underlying structure even though we initialize the Markov chains randomly in the full square $[-1,1]\times [-1, 1]$. This indicates again that the score function derived from the learned density is useful in sparse regions and not only near the training data. \Cref{tab:regOTdist-moons} tabulates the regularized OT distance between a test set, consisting of realizations not used for training, and samples generated using the proposed approach and diffusion model. \Cref{fig:moons-diffusion} shows the samples generated using the diffusion model. 
\begin{table}[!b]
 \begin{minipage}[c]{0.60\textwidth}
  \centering
    \caption{Regularized OT distance between the generated samples and the test data for the two moons dataset}
    \label{tab:regOTdist-moons}
    \begin{tabular}{lcc}
    \toprule[1.5pt]
    Method & Proposed approach & Diffusion model \\
    \midrule[1pt]
    OT distance & 0.131 & 0.098 \\
    \bottomrule[1.5pt]
    \end{tabular}
 \end{minipage}
 \hfill
 \begin{minipage}[c]{0.35\textwidth}
  \centering
  \includegraphics[height=1.6in]{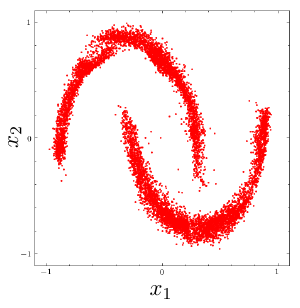}
    \captionof{figure}{Samples generated using a diffusion model trained on the two moons dataset}
    \label{fig:moons-diffusion}
 \end{minipage}
\end{table}

\begin{table}[!t]
	\centering
	
\end{table}

\subsubsection{Checkerboard}

\begin{figure}[p]
	\centering
	\includegraphics[height=0.9\textheight]{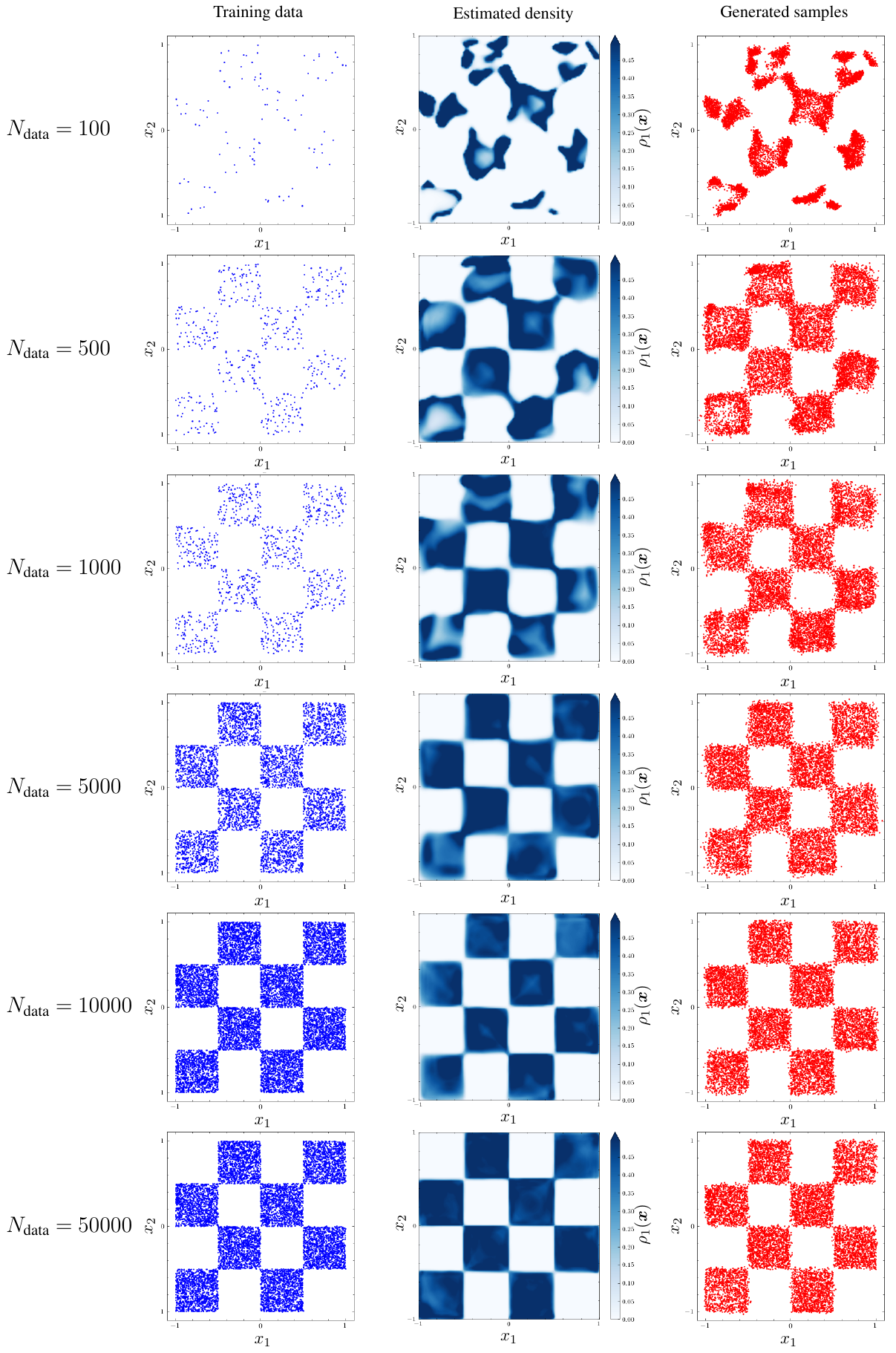}
	\caption{Training data, density estimated using the proposed approach and new samples for the checkerboard dataset for different number $N_{\mathrm{data}}$ of training data}
	\label{fig:checkerboard-results}
\end{figure}
\begin{figure}[t]
    \centering
    \includegraphics[width=0.75\linewidth]{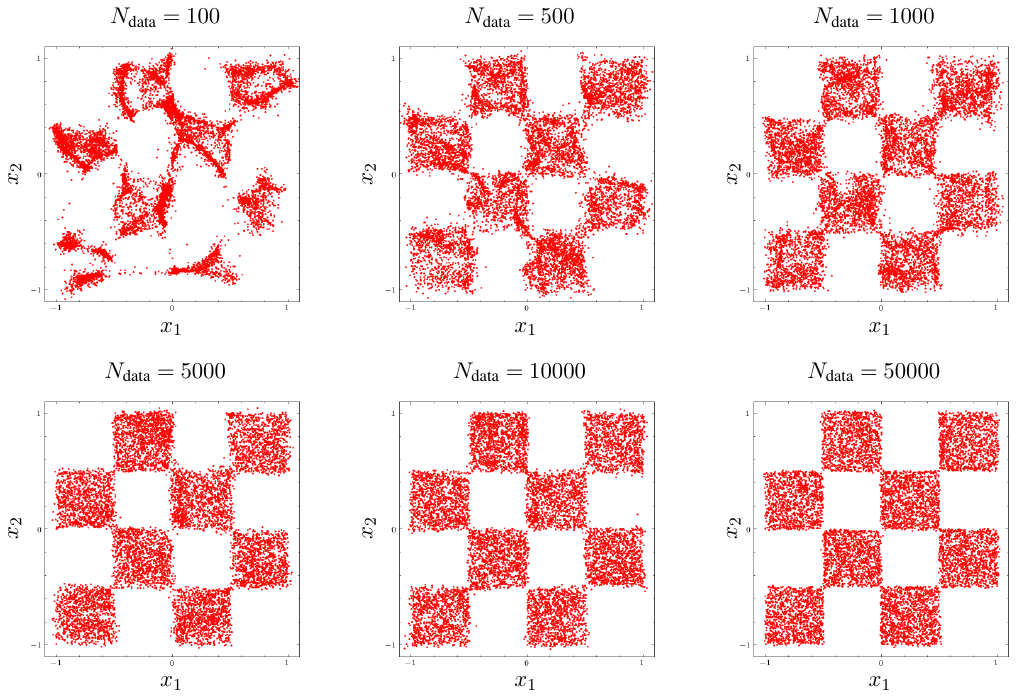}
    \caption{Samples generated using a diffusion model for the checkerboard data with varying amounts of training data $N_{\mathrm{data}}$}
    \label{fig:checkerboard-diffusion-samples}
\end{figure}

This dataset is sampled from the following data-generating distribution
\begin{equation}\label{eq:checkerboard}
	X_1 = 2U_1 - 1 , X_2 = 0.5 \left\{ p U_2 + \bmod (\lfloor X_1 \rfloor, 2) \right\},
\end{equation}
where $U_1, U_2 \sim \mathcal{U}(0, 1)$, $p \sim \text{Bern}(0.5)$, $\bmod(\cdot, \cdot)$ denotes the modulo operator and $\lfloor \cdot \rfloor$ denotes the floor function. The first column of \Cref{fig:checkerboard-results} shows the training data corresponding to different values of $N_{\mathrm{data}}$. Using the training data, we train a time-dependent classifier with $N = 10$ time steps chosen uniformly on a logarithmic scale. The second column in \Cref{fig:checkerboard-results} shows the estimated density. The third column in \Cref{fig:checkerboard-results} shows 10,000 new samples generated using ULA with the score function derived from the estimated density, 200 steps, and step size of 0.0001. As in the previous experiments, we seed individual Markov chains with realizations of a random variable uniformly distributed over $[-1, 1] \times [-1, 1]$. In line with expectations, the estimated density and the generated realizations qualitatively improve as more training data becomes available. \Cref{tab:regOTdist-checkerboard} reports the regularized OT distance between 10,000 samples generated using the proposed approach and diffusion models, from 10,000 test realizations. \Cref{tab:regOTdist-checkerboard} shows quantitatively that the generated realizations are similar beyond $N_{\mathrm{data}} = 500$. In comparison to \Cref{fig:checkerboard-results}, \Cref{fig:checkerboard-diffusion-samples} shows the realizations generated using the diffusion models for various amounts of training data $N_{\mathrm{data}}$. 
\begin{table}[H]
	\centering
	\caption{Regularized OT distance between the generated samples and the test data for the checkerboard data}
	\label{tab:regOTdist-checkerboard}
	\begin{tabular}{lcccccc}
	\toprule[1.5pt]
        & \multicolumn{6}{c}{$N_{\mathrm{data}}$} \\
        \cline{2-7}
	Method & 100 & 500 & 1,000 & 5,000 & 10,000 & 50,000 \\
        \midrule[1pt]
        Proposed approach & 0.022 & 0.016 & 0.017 & 0.014 & 0.013 & 0.015 \\
        Diffusion model & 0.152 & 0.070 & 0.058 & 0.049 & 0.040 & 0.036 \\
	\bottomrule[1.5pt]
	\end{tabular}
\end{table}

\begin{figure}[!h]
	\centering
	\includegraphics{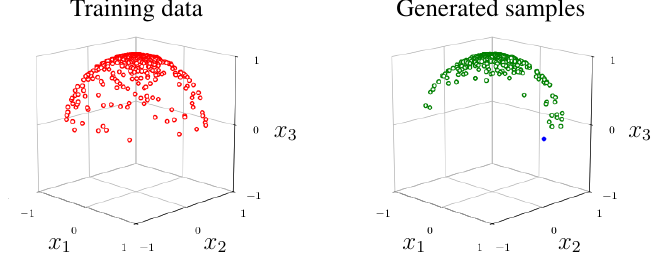}
	\caption{Training data (left) and generated samples (right) for the 3-dimensional semisphere. We show 500 generated samples only. The blue dot (\textcolor{blue}{$\bullet$}) among the generated samples denotes the initial seed $[1, 0, 0]\transpose$ for the Markov chains}
    \label{fig:hemisphere3-training-generated-data}
\end{figure}

\subsubsection{Multidimensional manifold}
\begin{figure}[!ht]
	\centering
	\includegraphics[width=0.9\textwidth]{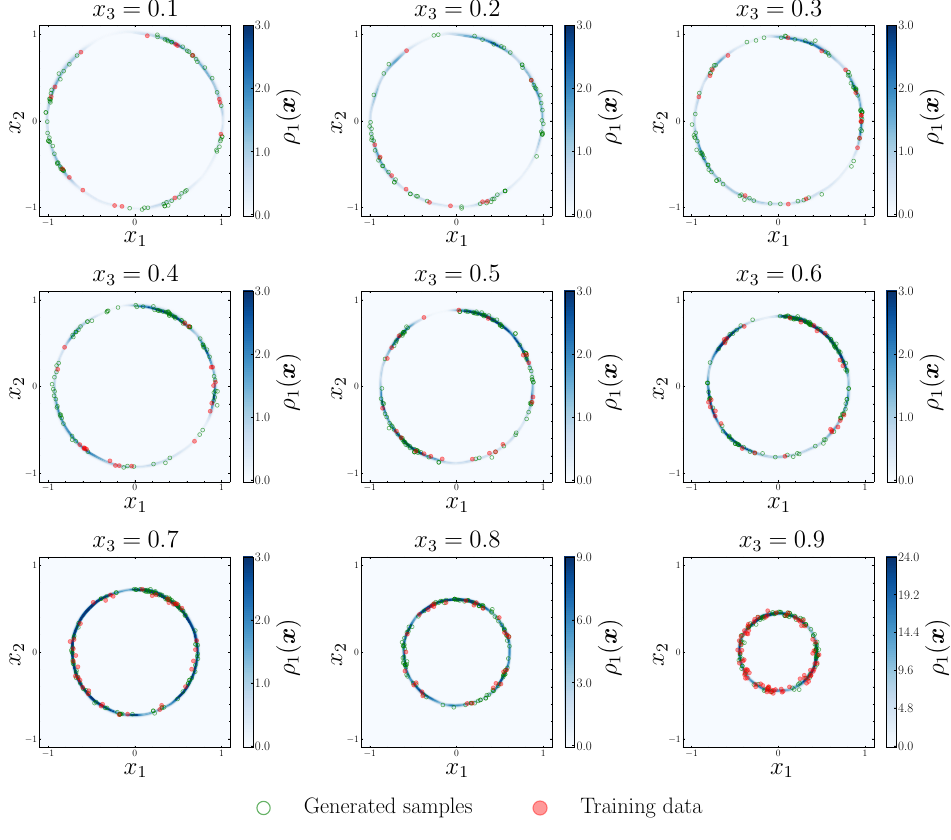}
	\caption{Density estimated using the proposed approach for a 3-dimensional semisphere at different values of $\x_3$. We also show the training data (\textcolor{red!40}{$\bullet$}) and a few generated realizations (\textcolor{green!70!black!80}{$\circ$}) around the same value of $x_3$}
	\label{fig:hemisphere3-density-learned}
	\vspace{1em}
%
%
	\includegraphics[width=\textwidth]{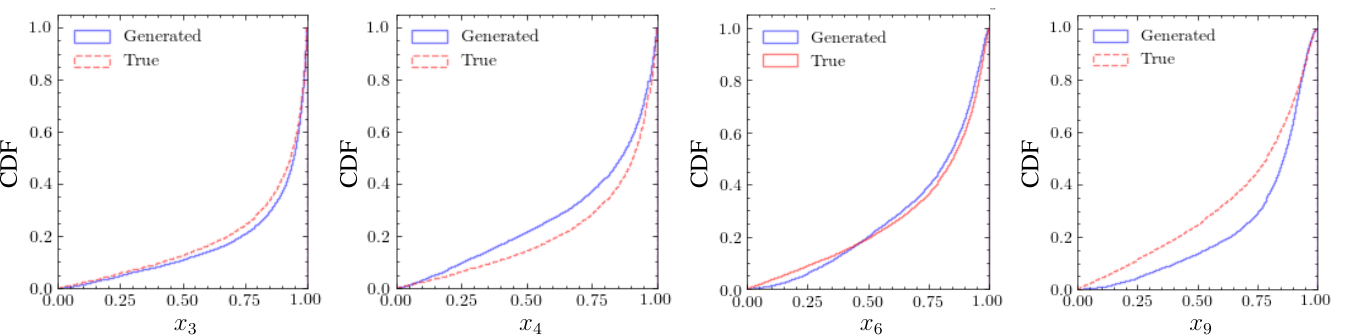}
	\caption{Empirical marginal cumulative distribution functions for $x_n$ estimated from the test data and generated samples for $n$-dimensional semisphere, where $n = \left\{ 3,4,6,9\right\}$ from left to right}
	\label{fig:hemisphere-cdf-compare}
\end{figure}

In this example, which we adapt from \cite{gottwald2024stable}, we demonstrate that the proposed approach can perform density estimation over low-dimensional manifolds in higher-dimensional spaces. We generate the training data by first sampling $\Z = [Z_1, Z_2, \ldots, Z_{n}]\transpose$ from a $n$-dimensional standard normal variable, and then set $\Y = [Z_1, Z_2, \ldots, \alpha \lvert Z_n\rvert]$, where $\alpha = 5$ and $\lvert \cdot \rvert$ denotes the absolute value. Subsequently, we compute $\X = (1 + \beta) \Y / \lVert \Y \rVert_2$, where $\beta \sim \mathcal{U}(0, 0.01)$ is a perturbation in the radial direction. We consider $n = \{ 3, 4, 6, 9\}$ and the training dataset $\mathcal{D}$ consists of $N_{\mathrm{data}} = 1000$ data points in each case. \Cref{fig:hemisphere3-training-generated-data} shows the training data in dimension $n = 3$. The training data is used to train the time-dependent classifier with $N = 20$ time instants sampled uniformly on a logarithmic scale.  Next to the training data in \Cref{fig:hemisphere3-training-generated-data} are new realizations of $\X$ sampled using the trained classifier. For all values of $n$, we generate 10,000 new samples using the No U-Turn HMC algorithm~\cite{hoffman2014no} with a step size of 0.001, 100 leap-frog steps, a target acceptance rate of 0.75, and discard the first 500 steps to allow for burn-in.  We use the \texttt{hamiltorch} package~\cite{cobb2023hamiltorch} for sampling and use $[1, 0, 0]\transpose$ as the initial condition. \Cref{fig:hemisphere3-density-learned} shows the density estimated using the proposed approach for different fixed values of $\x_3$. We also show the training data and a few generated realizations in the neighborhood of the corresponding $\x_3$ plane for reference in \Cref{fig:hemisphere3-density-learned}. \Cref{fig:hemisphere3-training-generated-data,fig:hemisphere3-density-learned} show that the proposed approach is able to detect the low-dimensional manifold over which the probability density concentrates and faithfully generates new samples that reside on it. 

Next, we assess the quality of the generated samples. \Cref{fig:hemisphere-cdf-compare} compares the empirical marginal cumulative distribution functions (cdf) for the last variable $x_{n}$ estimated using 10,000 test data points, not used for training, and the generated realizations. The cdf estimated from the generated samples matches the cdf estimated from the test data points except when $n=9$. We also compute the regularized OT distance between the generated samples and the test set using the Sinkhorn-Knopp algorithm~\cite{cuturi2013sinkhorn} with the regularization parameter set to 0.01. \Cref{tab:hemisphere-sinkhorn-distance} tabulates the regularized OT distance for different values of $n$. For comparison, \Cref{tab:hemisphere-sinkhorn-distance} also shows the regularized OT distance between samples generated using a diffusion model and the test set. \Cref{tab:hemisphere-sinkhorn-distance} shows that the proposed approach is marginally better than a diffusion model. However, both the empirical CDF and the OT distance suggest that the proposed approach and diffusion models struggle to sample from the manifold when $n=9$. This is due to the challenging nature of the task, which stems from the probability mass increasingly concentrating near the cap of the hyper-semisphere, a region that becomes exponentially smaller with increasing dimensionality $n$ of the ambient space.
\begin{table}[t]
	\centering
	\caption{Regularized OT distance between the generated samples and the test data for the multi-dimensional manifold}
	\label{tab:hemisphere-sinkhorn-distance}
	\begin{tabular}{ccc}
	\toprule[1.5pt]
        Dimension $n$ &  Proposed approach & Diffusion model \\
	\midrule
	3	&	0.021 &  0.021\\
	4   &	0.031 &  0.121\\
	6   &	0.238 &  0.260\\
	9   &	0.389 &  0.416\\
	\bottomrule[1.5pt]
	\end{tabular}
\end{table}

\subsubsection{20-dimensional dataset}
\begin{figure}[!t]
    \centering
    \includegraphics[width=0.95\textwidth]{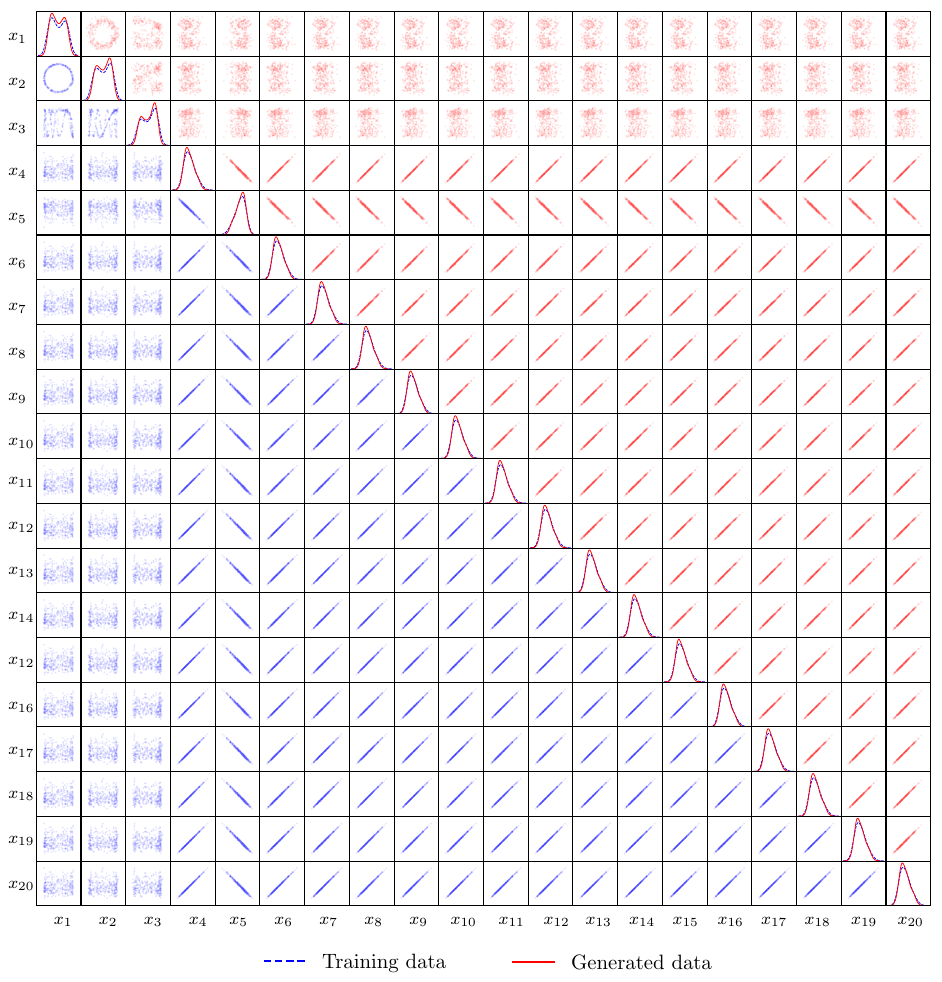}
    \caption{Training data and samples generated using the proposed approach for the 20-dimensional example. The joint realizations between pairs of random variables from the training data and generated samples are shown in the lower and upper off-diagonal plots, respectively. The diagonal plots show the kernel density estimate of the marginal distribution estimated from the generated samples (\textcolor{red}{---}) and training data (\textcolor{blue}{$--$})}
    \label{fig:20D-data}
\end{figure}

We consider a still higher dimensional test case where the dataset consists of $N_{\mathrm{data}} = 300$ realizations of a 20-dimensional random vector (\ie $n=20$), consisting of some independent and linearly correlated component random variables. The data generation process and the density $\rho_{\mathrm{data}}$ associated with the data generating distribution are unknown, but the training data is publicly available~\footnote{  \url{https://github.com/sanjayg0/PLoM/blob/main/example/example0/data/data_example20D.csv}}. So, the goal is to generate new samples that are statistically similar to the available training data. We use this data to train the time-dependent classifier with $N=20$ time instants chosen uniformly on a logarithmic scale. \Cref{fig:20D-data} shows the training data: the lower off-diagonal plots show the joint data between pairs of random variables, and the diagonal plots show the kernel density estimate of the marginal distribution. \Cref{fig:20D-data} shows that some variables are near perfectly positively (\eg $X_4$ and $X_6$) or (\eg $X_4$ and $X_5$) negatively correlated with another variable. 

\begin{figure}[t]
	\centering
	\includegraphics[width=2in]{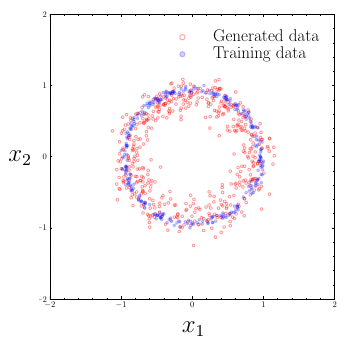}
	\caption{Realizations of $X_1$ and $X_2$ sampled from the density estimated using the proposed approach for the 20-dimensional dataset}
	\label{fig:20D-generated-samples}
\end{figure}

The plots in the upper off-diagonal in \Cref{fig:20D-data} show the joint realizations between pairs of random variables generated using the proposed approach. We sample 1,000 realizations using ULA with a step size of $10^{-5}$ and 1000 steps. We seed individual Markov chains with slightly perturbed points from the training data. More specifically, the initial seed of the Markov chain is an independently sampled realization from the training data set plus a zero-mean Gaussian perturbation with a standard deviation equal to 0.01. The diagonal plots in \Cref{fig:20D-data}  show the kernel density estimate of the marginal distribution obtained using the training data and the generated sample. Since the proposed approach yields the joint density between all 20 random variables, the marginals must be extracted using either Monte Carlo sampling or via kernel density estimation using the generated realizations. We adopt the latter approach to enable direct comparison. \Cref{fig:20D-generated-samples} is a close-up of the plots in the second row and first column and first row and second column of \Cref{fig:20D-data}, showing the generated realizations of $X_1$ and $X_2$ and the corresponding training data. The generated realizations are slightly more diffuse compared to the training data but evenly spread over the entire ring. Overall, \Cref{fig:20D-data,fig:20D-generated-samples} show that the proposed approach generates new realizations statistically similar to the training data, detecting any linear and non-linear correlation structure underlying the training data. The regularized OT distance between 10,000 samples generated using the proposed approach and 10,000 test realizations is 0.346. The regularized OT distance between samples generated using a diffusion model and the test data is 0.406. 

\subsubsection{Discussion}
Across all 2-dimensional examples, the proposed method effectively captures any disconnected underlying structure in the data-generating density. The learned densities match the empirical distribution well, but some artefacts arise in regions with sparse training data. In higher-dimensional problems, the method demonstrates strong performance: it faithfully generates data concentrated along low-dimensional manifolds embedded in high-dimensional ambient spaces. Although we observe some degradation in sampling quality as the ambient dimension increases, as the data concentrates in exponentially small regions, the method continues to generate samples that respect the underlying structure. The proposed approach also preserves the correlation structures. These results highlight the efficacy of the proposed approach in time-independent or static density estimation, even with simple stochastic interpolants and in the low-data regime.



\subsection{Application of the proposed approach to detecting rare events}\label{subsec:outlier-detection-results}

The previous section shows that the proposed approach is useful for static density estimation using a training dataset consisting of iid realizations from the data-generating distribution with density $\rho_{\mathrm{data}}$. An important application of density estimation is rare event detection, whereby samples in the low-density region are considered rare.  Formally, we describe the rare event detection task as follows. Given a dataset $\mathcal{D} = \left\{ \x^{(1)}, \x^{(2)}, \ldots, \x^{(N_{\mathrm{data}})} \right\}$, the goal is to assign an rarity score $m(\x) = -\log\rho_1(\x)$, such that a larger score means that the corresponding data point $\x$ is relatively far away from the modes of $\rho_{\mathrm{data}}$. The approach is unsupervised because we do not use any information from labels. \Cref{sec:generative-model} describes how we estimate $\rho_{\mathrm{data}}$ from $\mathcal{D}$. In subsequent examples, we demonstrate the efficacy of our approach on rare event detection tasks on image datasets from the ADBench suite~\cite{han2022adbench}, which provides the most comprehensive benchmark datasets for this purpose.


In this study, we consider three dataset collections from the ADBench suite~\cite{han2022adbench} --- CIFAR10~\cite{krizhevsky2009learning}, Fashion MNIST~\cite{xiao2017} and MNIST-C~\cite{mu2019mnist}. The CIFAR10 collection consists of 10 different datasets: each dataset is created by sampling 5,000 images from one of the multi-classes and treating them as common events, and randomly sampling 263 images uniformly across the remaining classes and treating them as rare. The FashionMNIST (FMNIST) collection also consists of 10 different datasets: each dataset is created by sampling 6,000 images from one of the multi-classes and treating them as common, and sampling 315 images uniformly across the remaining classes and treating them as rare. Finally, the MNIST-C collection consists of 15 datasets: each dataset is created by sampling 9,500 uncorrupted images from the several classes of the MNIST data and treating them as common, and sampling 500 corrupted MNIST images and treating them as rare. On average, each dataset is designed to have approximately 5\% outliers. We refer interested readers to Appendix B2 and Table B1 in \cite{han2022adbench} for more details regarding this dataset. 

\begin{table}[!b]
 \begin{minipage}[c]{0.45\textwidth}
  \centering
  \includegraphics[height=3in]{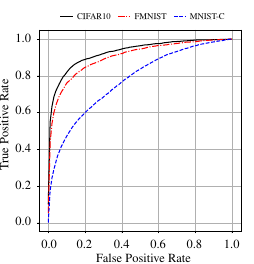}
    \captionof{figure}{Average Receiver Operating Characteristic (ROC) curve of the rare-event detector based on the proposed density estimation approach for CIFAR10, FMNIST and MNIST-C collections}
    \label{fig:ROC-curves}
 \end{minipage}
 \hfill
 \begin{minipage}[c]{0.45\textwidth}
  \centering
  \caption{Average area under the ROC curve (AUC-ROC) on various benchmark datasets for our method and ODIM~\cite{kim2023odim}. Bold values indicate the best performance}
	\label{tab:AUC-ROC}
	\begin{tabular}{lcc}
	\toprule[1.5pt]
	\multirow{2}{*}{\makecell{Collection}} & \multicolumn{2}{c}{AUC-ROC}\\
        \cline{2-3}
         & our method & ODIM \\
	\midrule
	CIFAR10	&	\textbf{0.933} & 0.922\\
	FMNIST   &	\textbf{0.909} & \textbf{0.909}\\
	MNIST-C   &	\textbf{0.775} & 0.736\\
        \bottomrule[1.5pt]
	\end{tabular}
 \end{minipage}
\end{table}
In these experiments, the network $f$ is the composition of a pre-trained feature extractor $f_2$ and a learnable time-dependent classification head $f_1$, \ie $f_{\thetaa} = f_{1, \thetaa} \circ f_2$. Following \cite{han2022adbench}, we use a pre-trained ViT as the feature extractor, which yields a 1000-dimensional feature vector for every image. These embeddings are provided within ADBench. Therefore, each point in the training dataset $\mathcal{D}$ is a 1000-dimensional vector, \ie $n=1000$ in these experiments. We train the time-dependent classifier on these datasets, using stochastic interpolants to induce a stochastic process between $t=0$ and $t=1$. \Cref{appsubsec:outlier-models} provides additional details on the architecture of $f$ and learning hyper-parameters for each dataset. After training, we evaluate $\log \rho_1$ and use it to sort the training data points in increasing value of $\log \rho_1$ such that images with smaller values of $\log \rho_1$ are considered rare. \Cref{fig:ROC-curves} shows the receiver operating characteristic (ROC) curve from the resulting rank-ordered list for the three collections, averaged across all the datasets within each collection. We report the area under the average area under the ROC curve (AUC-ROC) in \Cref{tab:AUC-ROC} and compare the results against ODIM~\cite{kim2023odim} --- a recently proposed state-of-the-art method that uses an under-fitted generative model to detect rare events. We collect the performance metrics for ODIM from Table~C.4.2 in \cite{kim2023odim}. \Cref{tab:AUC-ROC} shows that our approach outperforms ODIM on CIFAR10 and MNIST-C collections, and is competitive on the FMNIST collection. \Cref{fig:CIFAR10-outliers,fig:FMNIST-outliers} in \ref{appsubsec:outlier-detection-results} shows the top fifty entries in the rank-ordered list of images for the best and worst performing class within the CIFAR10 and FMNIST collections, respectively. We were unable to determine the image corresponding to the embeddings for the MNIST-C datasets.  Collectively, these results show that the proposed approach is also useful as a rare event detection tool. 

\section{Conclusions and outlook}\label{sec:conclusion}

In this work, we propose an explicit method to learn the temporal evolution of probability densities from sample path observations using a time-dependent binary classifier. Central to our approach is the construction of a classifier whose architecture incorporates the time separation between samples, enabling it to learn the partial derivative of the evolving log-density using a contrastive estimation-based objective. The resulting method is fully data-driven, does not require knowledge of the dynamics of the underlying stochastic processes, and scales naturally to high-dimensional problems. Through examples involving stochastic dynamical systems, density estimation and sampling, and rare event detection, we show the efficacy and versatility of the proposed approach. In particular, the proposed approach delivers competitive or superior performance compared to state-of-the-art methods on generative modeling and rare event detection. Beyond its immediate applications, the flexibility of the proposed method opens up several future directions for research. Among them, we aim to extend the approach to conditional density estimation, enabling its use in Bayesian inference and experimental design. Additionally, coupling the method with manifold learning or dimension reduction techniques may significantly broaden its applicability to high-dimensional problems. 
 
\section{Acknowledgments}
	
The authors acknowledge support from ARO grant W911NF2410401. The authors also acknowledge the Center for Advanced Research Computing (CARC, \href{https://carc.usc.edu}{carc.usc.edu}) at the University of Southern California for providing computing resources that have contributed to the research results reported within this publication.

\appendix
\setcounter{section}{0}
\renewcommand{\appendixname}{Appendix}
\renewcommand{\thesection}{Appendix~\Alph{section}}
\renewcommand{\thesubsection}{\Alph{section}\arabic{subsection}}
\renewcommand{\thefigure}{\Alph{section}\arabic{figure}}
\setcounter{figure}{0}
\renewcommand{\thetable}{\Alph{section}\arabic{table}}
\setcounter{table}{0}

\section{Additional details regarding various experiments}

\subsection{Additional details for experiments in \Cref{subsec:results-javier}}\label{appsubsec:time-dependent-experiments}

This appendix provides additional details on the setup for experiments in \Cref{subsec:results-javier}. The time variable is embedded using 16-dimensional random Fourier features~\cite{tancik2020fourier} and concatenated along the spatial coordinates. The network $f_{\thetaa}$ consists of 3 hidden layers with 512 neurons each and \texttt{SiLU} activation for both problems. In all the experiments in \Cref{subsec:results-javier}, we minimize \Cref{eq:loss-MCS} to train the network $f_{\thetaa}$ using the Adam optimizer for 10,000 epochs with batch size $N_{\mathrm{b}} = $ 100, weight decay $10^{-5}$,  and learning rate equal to 0.001 with an exponential scheduler ($\beta=0.9997$). 
        
\subsection{Additional details for experiments in \Cref{sec:results-dens-est-sampling}}\label{appsubsec:models}

This appendix provides additional details related to the density estimation experiments from \Cref{sec:results-dens-est-sampling}. The network $f_{\thetaa}$ for the circles, moons, and checkerboard data consists of 3 hidden layers with 256 neurons each and \texttt{ReLU} activation. For these datasets, we simply append the time variable $t$ with the spatial inputs. For the multi-dimensional manifold and 20-dimensional examples, $f_{\thetaa}$ consists of 4 hidden layers with 256 neurons each and \texttt{ReLU} activation. For these datasets, however, we use a 16-dimensional Fourier feature-based embedding for time. We append these embeddings with the spatial inputs. In all the experiments in \Cref{sec:results-dens-est-sampling}, we minimize \Cref{eq:loss-MCS} to train the network $f_{\thetaa}$ using the Adam optimizer with batch size $N_{\mathrm{b}} = $ 5,000 and learning rate equal to 0.001. 
We oversample the training dataset if the batch size $N_{\mathrm{b}}$ is greater than the number of training data points. We train the classifier for 5,000 epochs for the circles, moons, and checkerboard datasets. We train the classifier for 20,000 and 2,000 epochs on the multi-dimensional manifold and 20-dimensional datasets, respectively. 

We refer interested readers to \cite{dasgupta2025unifying} for a background on diffusion models. We use the variance preserving variant of diffusion models for all the experiments in \Cref{sec:results-dens-est-sampling}. In this setting, the equivalent to \Cref{eq:linear-interpolant} is
\begin{equation}
    \x_{t} = \sigma(t) \x_0 + m(t) \x \quad \forall t \in [0, 1],
\end{equation}
where the latent variable $\x_0 \sim \mathcal{N}(0, \mathbb{I}_n)$,  
\begin{equation}\label{eq:variance-preserving-schedule}
     m(t) =  \exp \left( - \frac{1}{2} \int_0^t \beta(s) \,\mathrm{d}s \right), \text{ and } \sigma^2(t) = \left( 1 - m^2(t) \right).
\end{equation}
In \Cref{eq:variance-preserving-schedule}, we choose $\beta(t) = \beta_{\mathrm{min}} + t (\beta_{\mathrm{max}} - \beta_{\mathrm{min}})$, with $\beta_{\mathrm{min}} = 0.001$ and $\beta_{\mathrm{max}} = 5$. The score network is a fully connected network with 4 hidden layers of width 256 and \texttt{ReLU} activation. The time variable is embedded using a 16-dimensional random Fourier features and concatenated along the spatial coordinates as in \cite{baptista2025memorization}. The score network is trained for 100,000 epochs using the Adam optimizer with a batch size of 5,000 and the learning rate set to 0.001. Similar to \cite{dasgupta2025unifying}, we use an adaptive Explicit Runge-Kutta method of order 5(4), available through \texttt{SciPy}'s~\cite{2020SciPy-NMeth} \texttt{solve\textunderscore ivp} routine, to integrate the reverse time probability flow ODE and generate samples. Note that the score function had more parameters than the time-dependent classifier and had to be trained for a longer duration on all the examples.


\subsection{Additional details for experiments in \Cref{subsec:outlier-detection-results}}\label{appsubsec:outlier-models}

In this appendix, we provide details related to experiments in \Cref{subsec:outlier-detection-results}. The network for CIFAR10 and FMNIST has three hidden layers with 4800, 1200, and 300 neurons, respectively. For the MNIST-C datasets, we use the network that has six hidden layers with 4800, 4800, 4800, 4800, 1200, and 300 neurons, respectively. We use \texttt{ReLU} activation in these networks. To embed the time variable $t$, we map it to a $n_{\ell}$-dimensional vector using a shallow network with one hidden layer and \texttt{swish} activation function. We use this vector as an input for conditional instance normalization (CIN) operating on the activations after each hidden layer. $n_{\ell}$ is 20 for CIFAR10 and FMNIST, and 50 for MNIST-C. We minimize the loss function using the Adam optimizer. We report the batch size, learning rate, and the number of time steps $N$ in \Cref{tab:OD-hyperparameters}. These time steps are chosen uniformly between $[0,1]$ on a linear scale. Before evaluating the classifier, we normalize the embedding for each image by its Euclidean norm (except for the brightness corruption dataset of MNIST-C).
\begin{table}[t]
	\centering
	\caption{Learning-related hyper-parameters for the rare event detection experiments in \Cref{subsec:outlier-detection-results}. LR and $N_\mathrm{b}$ denote the learning rate and batch size, respectively}
	\label{tab:OD-hyperparameters}
	\begin{tabular}{lcccc}
		\toprule[1.5pt]
		\multirow{2}{*}{\makecell{Dataset\\ collection}} &  \multicolumn{4}{c}{Hyper-parameters} \\
		\cline{2-5}
		 & Epoch & LR & $N_{\mathrm{b}}$ & $N$ \\
		\midrule
		CIFAR10 & 4000 & $10^{-5}$ & 1350 & 50\\
		FMNIST  & 4000 & $10^{-5}$ & 1300 & 50\\
		MNIST-C & 2000 & $10^{-5}$ & 1200 & 50\\
		\bottomrule[1.5pt]
	\end{tabular}
\end{table}	

\section{Top fifty outliers corresponding to results in \Cref{subsec:outlier-detection-results}}\label{appsubsec:outlier-detection-results}

\begin{figure}
    \centering
    \includegraphics[width=\linewidth]{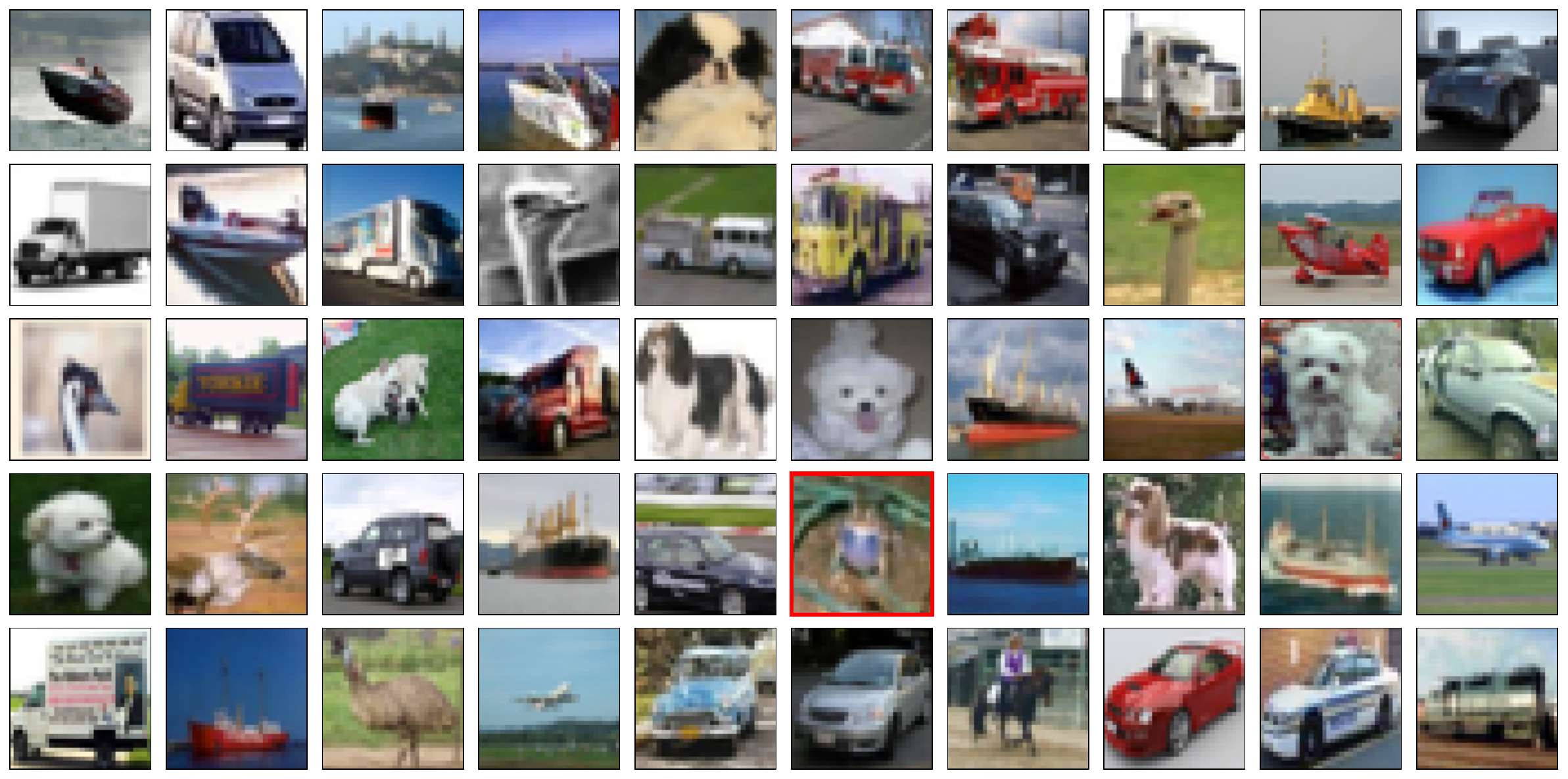}\\
    \vspace{2em}
    \includegraphics[width=\linewidth]{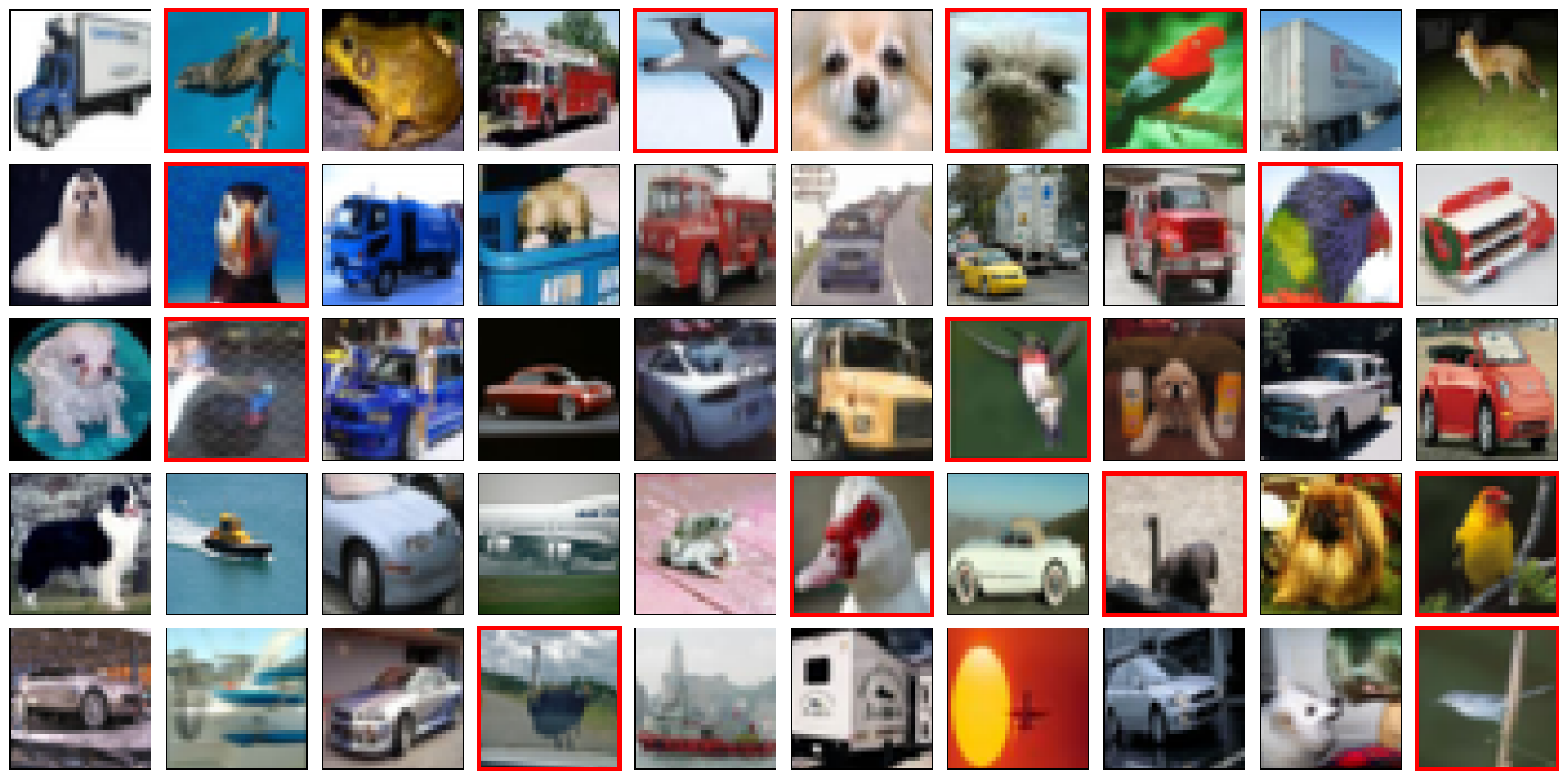}
    \caption{Top fifty outliers from the dataset that is part of the CIFAR10 collection and consists mostly of frog (top) and bird (bottom) images}
    \label{fig:CIFAR10-outliers}
\end{figure}

\begin{figure}
    \centering
    \includegraphics[width=\linewidth]{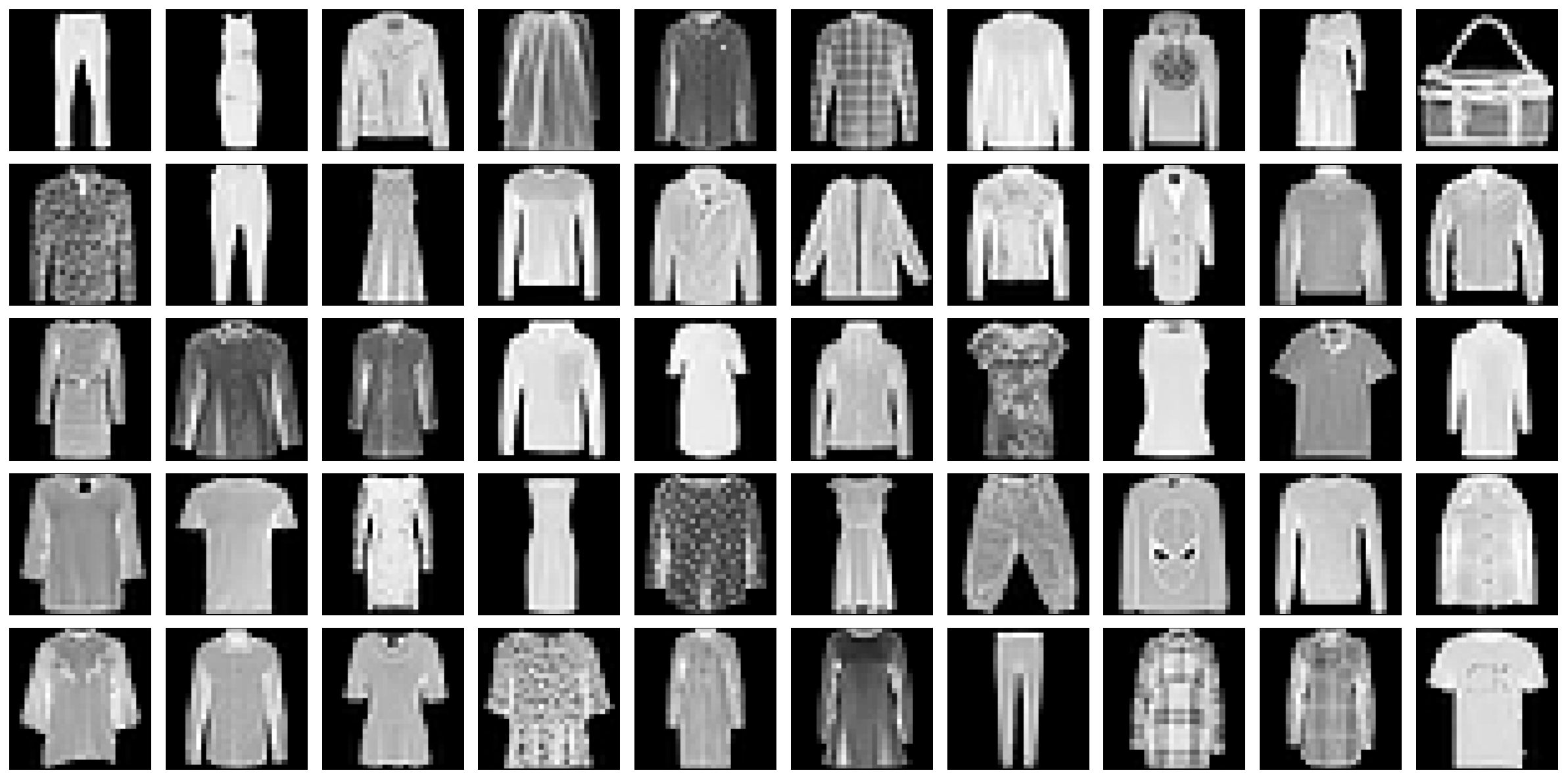}\\
    \vspace{2em}
    \includegraphics[width=\linewidth]{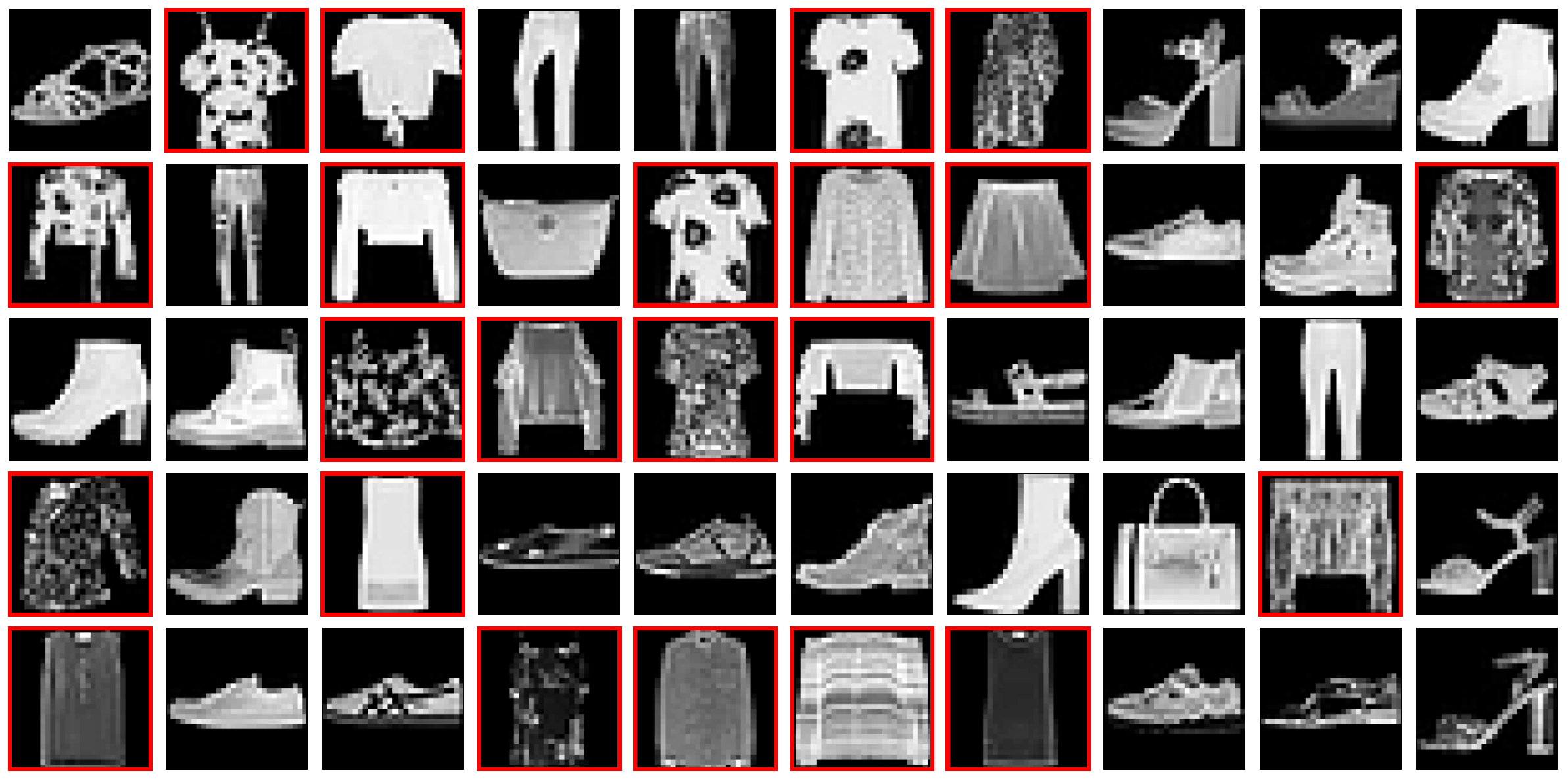}
    \caption{Top fifty outliers from the dataset that is part of the FMNIST collection and consists mostly of images of sneakers (top) and shirts (bottom)}
    \label{fig:FMNIST-outliers}
\end{figure}

For the CIFAR10 and FMNIST collections, we present the top fifty outliers for the datasets on which our method performs the best and worst in \Cref{fig:CIFAR10-outliers} and \Cref{fig:FMNIST-outliers}, respectively. We mark images from the normal class with a red border around them. 


\bibliographystyle{elsarticle-num-names} 
\bibliography{references}	

\begin{thebibliography}{76}
\expandafter\ifx\csname natexlab\endcsname\relax\def\natexlab#1{#1}\fi
\providecommand{\url}[1]{\texttt{#1}}
\providecommand{\href}[2]{#2}
\providecommand{\path}[1]{#1}
\providecommand{\DOIprefix}{doi:}
\providecommand{\ArXivprefix}{arXiv:}
\providecommand{\URLprefix}{URL: }
\providecommand{\Pubmedprefix}{pmid:}
\providecommand{\doi}[1]{\href{http://dx.doi.org/#1}{\path{#1}}}
\providecommand{\Pubmed}[1]{\href{pmid:#1}{\path{#1}}}
\providecommand{\bibinfo}[2]{#2}
\ifx\xfnm\relax \def\xfnm[#1]{\unskip,\space#1}\fi
\bibitem[{Han et~al.(2018)Han, Jentzen, and E}]{han2018solving}
\bibinfo{author}{J.~Han}, \bibinfo{author}{A.~Jentzen}, \bibinfo{author}{W.~E},
\newblock \bibinfo{title}{Solving high-dimensional partial differential
  equations using deep learning},
\newblock \bibinfo{journal}{Proceedings of the National Academy of Sciences}
  \bibinfo{volume}{115} (\bibinfo{year}{2018}) \bibinfo{pages}{8505--8510}.
\bibitem[{Cho et~al.(2016)Cho, Venturi, and Karniadakis}]{cho2016numerical}
\bibinfo{author}{H.~Cho}, \bibinfo{author}{D.~Venturi}, \bibinfo{author}{G.~E.
  Karniadakis},
\newblock \bibinfo{title}{Numerical methods for high-dimensional probability
  density function equations},
\newblock \bibinfo{journal}{Journal of Computational Physics}
  \bibinfo{volume}{305} (\bibinfo{year}{2016}) \bibinfo{pages}{817--837}.
\bibitem[{Soize(2023)}]{soize2023overview}
\bibinfo{author}{C.~Soize},
\newblock \bibinfo{title}{An overview on uncertainty quantification and
  probabilistic learning on manifolds in multiscale mechanics of materials},
\newblock \bibinfo{journal}{Mathematics and Mechanics of Complex Systems}
  \bibinfo{volume}{11} (\bibinfo{year}{2023}) \bibinfo{pages}{87--174}.
\bibitem[{Soize and Ghanem(2024)}]{soize2024probabilistic}
\bibinfo{author}{C.~Soize}, \bibinfo{author}{R.~Ghanem},
\newblock \bibinfo{title}{Probabilistic-learning-based stochastic surrogate
  model from small incomplete datasets for nonlinear dynamical systems},
\newblock \bibinfo{journal}{Computer Methods in Applied Mechanics and
  Engineering} \bibinfo{volume}{418} (\bibinfo{year}{2024})
  \bibinfo{pages}{116498}.
\bibitem[{Zhong et~al.(2023)Zhong, Navarro, Govindjee, and
  Deierlein}]{zhong2023surrogate}
\bibinfo{author}{K.~Zhong}, \bibinfo{author}{J.~G. Navarro},
  \bibinfo{author}{S.~Govindjee}, \bibinfo{author}{G.~G. Deierlein},
\newblock \bibinfo{title}{Surrogate modeling of structural seismic response
  using probabilistic learning on manifolds},
\newblock \bibinfo{journal}{Earthquake Engineering \& Structural Dynamics}
  \bibinfo{volume}{52} (\bibinfo{year}{2023}) \bibinfo{pages}{2407--2428}.
\bibitem[{El~Moselhy and Marzouk(2012)}]{el2012bayesian}
\bibinfo{author}{T.~A. El~Moselhy}, \bibinfo{author}{Y.~M. Marzouk},
\newblock \bibinfo{title}{Bayesian inference with optimal maps},
\newblock \bibinfo{journal}{Journal of Computational Physics}
  \bibinfo{volume}{231} (\bibinfo{year}{2012}) \bibinfo{pages}{7815--7850}.
\bibitem[{Dasgupta et~al.(2025)Dasgupta, Ramaswamy, Murgoitio-Esandi, Foo, Li,
  Zhou, Kennedy, and Oberai}]{dasgupta2025conditional}
\bibinfo{author}{A.~Dasgupta}, \bibinfo{author}{H.~Ramaswamy},
  \bibinfo{author}{J.~Murgoitio-Esandi}, \bibinfo{author}{K.~Y. Foo},
  \bibinfo{author}{R.~Li}, \bibinfo{author}{Q.~Zhou}, \bibinfo{author}{B.~F.
  Kennedy}, \bibinfo{author}{A.~A. Oberai},
\newblock \bibinfo{title}{Conditional score-based diffusion models for solving
  inverse elasticity problems},
\newblock \bibinfo{journal}{Computer Methods in Applied Mechanics and
  Engineering} \bibinfo{volume}{433} (\bibinfo{year}{2025})
  \bibinfo{pages}{117425}.
\bibitem[{Dasgupta and Johnson(2024)}]{dasgupta2024rein}
\bibinfo{author}{A.~Dasgupta}, \bibinfo{author}{E.~A. Johnson},
\newblock \bibinfo{title}{{REIN}: {R}eliability estimation via importance
  sampling with normalizing flows},
\newblock \bibinfo{journal}{Reliability Engineering \& System Safety}
  \bibinfo{volume}{242} (\bibinfo{year}{2024}) \bibinfo{pages}{109729}.
\bibitem[{Murphy(2022)}]{murphy2022probabilistic}
\bibinfo{author}{K.~P. Murphy}, \bibinfo{title}{Probabilistic machine learning:
  {A}n introduction}, \bibinfo{publisher}{MIT press}, \bibinfo{year}{2022}.
\bibitem[{Murphy(2023)}]{murphy2023probabilistic}
\bibinfo{author}{K.~P. Murphy}, \bibinfo{title}{Probabilistic machine learning:
  Advanced topics}, \bibinfo{publisher}{MIT press}, \bibinfo{year}{2023}.
\bibitem[{Sugiyama et~al.(2012)Sugiyama, Suzuki, and
  Kanamori}]{sugiyama2012density}
\bibinfo{author}{M.~Sugiyama}, \bibinfo{author}{T.~Suzuki},
  \bibinfo{author}{T.~Kanamori}, \bibinfo{title}{Density ratio estimation in
  machine learning}, \bibinfo{publisher}{Cambridge University Press},
  \bibinfo{year}{2012}.
\bibitem[{Garcia et~al.(2017)Garcia, Otero, Felix, Presedo, and
  Marquez}]{garcia2017nonparametric}
\bibinfo{author}{C.~A. Garcia}, \bibinfo{author}{A.~Otero},
  \bibinfo{author}{P.~Felix}, \bibinfo{author}{J.~Presedo},
  \bibinfo{author}{D.~G. Marquez},
\newblock \bibinfo{title}{Nonparametric estimation of stochastic differential
  equations with sparse {G}aussian processes},
\newblock \bibinfo{journal}{Physical Review E} \bibinfo{volume}{96}
  (\bibinfo{year}{2017}) \bibinfo{pages}{022104}.
\bibitem[{Ruttor et~al.(2013)Ruttor, Batz, and Opper}]{ruttor2013approximate}
\bibinfo{author}{A.~Ruttor}, \bibinfo{author}{P.~Batz},
  \bibinfo{author}{M.~Opper},
\newblock \bibinfo{title}{Approximate {G}aussian process inference for the
  drift function in stochastic differential equations},
\newblock \bibinfo{journal}{Advances in Neural Information Processing Systems}
  \bibinfo{volume}{26} (\bibinfo{year}{2013}).
\bibitem[{Gonz{\'a}lez-Garc{\'\i}a et~al.(1998)Gonz{\'a}lez-Garc{\'\i}a,
  Rico-Mart{\`\i}nez, and Kevrekidis}]{gonzalez1998identification}
\bibinfo{author}{R.~Gonz{\'a}lez-Garc{\'\i}a},
  \bibinfo{author}{R.~Rico-Mart{\`\i}nez}, \bibinfo{author}{I.~G. Kevrekidis},
\newblock \bibinfo{title}{Identification of distributed parameter systems: {A}
  neural net based approach},
\newblock \bibinfo{journal}{Computers \& chemical engineering}
  \bibinfo{volume}{22} (\bibinfo{year}{1998}) \bibinfo{pages}{S965--S968}.
\bibitem[{Brunton et~al.(2016)Brunton, Proctor, and
  Kutz}]{brunton2016discovering}
\bibinfo{author}{S.~L. Brunton}, \bibinfo{author}{J.~L. Proctor},
  \bibinfo{author}{J.~N. Kutz},
\newblock \bibinfo{title}{Discovering governing equations from data by sparse
  identification of nonlinear dynamical systems},
\newblock \bibinfo{journal}{Proceedings of the National Academy of Sciences}
  \bibinfo{volume}{113} (\bibinfo{year}{2016}) \bibinfo{pages}{3932--3937}.
\bibitem[{Riseth and Taylor-King(2017)}]{riseth2017operator}
\bibinfo{author}{A.~N. Riseth}, \bibinfo{author}{J.~P. Taylor-King},
\newblock \bibinfo{title}{Operator fitting for parameter estimation of
  stochastic differential equations},
\newblock \bibinfo{journal}{arXiv preprint arXiv:1709.05153}
  (\bibinfo{year}{2017}).
\bibitem[{Dai et~al.(2020)Dai, Gao, Lu, Zheng, and Duan}]{dai2020detecting}
\bibinfo{author}{M.~Dai}, \bibinfo{author}{T.~Gao}, \bibinfo{author}{Y.~Lu},
  \bibinfo{author}{Y.~Zheng}, \bibinfo{author}{J.~Duan},
\newblock \bibinfo{title}{Detecting the maximum likelihood transition path from
  data of stochastic dynamical systems},
\newblock \bibinfo{journal}{Chaos: An Interdisciplinary Journal of Nonlinear
  Science} \bibinfo{volume}{30} (\bibinfo{year}{2020}).
\bibitem[{Yang et~al.(2022)Yang, Daskalakis, and
  Karniadakis}]{yang2022generative}
\bibinfo{author}{L.~Yang}, \bibinfo{author}{C.~Daskalakis},
  \bibinfo{author}{G.~E. Karniadakis},
\newblock \bibinfo{title}{Generative ensemble regression: {L}earning particle
  dynamics from observations of ensembles with physics-informed deep generative
  models},
\newblock \bibinfo{journal}{SIAM Journal on Scientific Computing}
  \bibinfo{volume}{44} (\bibinfo{year}{2022}) \bibinfo{pages}{B80--B99}.
\bibitem[{Wasserman(2006)}]{wasserman2006all}
\bibinfo{author}{L.~Wasserman}, \bibinfo{title}{All of nonparametric
  statistics}, \bibinfo{publisher}{Springer Science \& Business Media},
  \bibinfo{year}{2006}.
\bibitem[{K{\"o}the(2023)}]{kothe2023review}
\bibinfo{author}{U.~K{\"o}the},
\newblock \bibinfo{title}{A review of change of variable formulas for
  generative modeling},
\newblock \bibinfo{journal}{arXiv preprint arXiv:2308.02652}
  (\bibinfo{year}{2023}).
\bibitem[{Yang et~al.(2020)Yang, Zhang, and Karniadakis}]{yang2020physics}
\bibinfo{author}{L.~Yang}, \bibinfo{author}{D.~Zhang}, \bibinfo{author}{G.~E.
  Karniadakis},
\newblock \bibinfo{title}{Physics-informed generative adversarial networks for
  stochastic differential equations},
\newblock \bibinfo{journal}{SIAM Journal on Scientific Computing}
  \bibinfo{volume}{42} (\bibinfo{year}{2020}) \bibinfo{pages}{A292--A317}.
\bibitem[{G{\"u}ler et~al.(2019)G{\"u}ler, Laignelet, and
  Parpas}]{guler2019towards}
\bibinfo{author}{B.~G{\"u}ler}, \bibinfo{author}{A.~Laignelet},
  \bibinfo{author}{P.~Parpas},
\newblock \bibinfo{title}{Towards robust and stable deep learning algorithms
  for forward backward stochastic differential equations},
\newblock \bibinfo{journal}{arXiv preprint arXiv:1910.11623}
  (\bibinfo{year}{2019}).
\bibitem[{Beck et~al.(2021)Beck, Becker, Grohs, Jaafari, and
  Jentzen}]{beck2021solving}
\bibinfo{author}{C.~Beck}, \bibinfo{author}{S.~Becker},
  \bibinfo{author}{P.~Grohs}, \bibinfo{author}{N.~Jaafari},
  \bibinfo{author}{A.~Jentzen},
\newblock \bibinfo{title}{Solving the {K}olmogorov {PDE} by means of deep
  learning},
\newblock \bibinfo{journal}{Journal of Scientific Computing}
  \bibinfo{volume}{88} (\bibinfo{year}{2021}) \bibinfo{pages}{1--28}.
\bibitem[{Chen et~al.(2021)Chen, Yang, Duan, and Karniadakis}]{chen2021solving}
\bibinfo{author}{X.~Chen}, \bibinfo{author}{L.~Yang},
  \bibinfo{author}{J.~Duan}, \bibinfo{author}{G.~E. Karniadakis},
\newblock \bibinfo{title}{Solving inverse stochastic problems from discrete
  particle observations using the {F}okker--{P}lanck equation and
  physics-informed neural networks},
\newblock \bibinfo{journal}{SIAM Journal on Scientific Computing}
  \bibinfo{volume}{43} (\bibinfo{year}{2021}) \bibinfo{pages}{B811--B830}.
\bibitem[{Lu et~al.(2022)Lu, Maulik, Gao, Dietrich, Kevrekidis, and
  Duan}]{lu2022learning}
\bibinfo{author}{Y.~Lu}, \bibinfo{author}{R.~Maulik}, \bibinfo{author}{T.~Gao},
  \bibinfo{author}{F.~Dietrich}, \bibinfo{author}{I.~G. Kevrekidis},
  \bibinfo{author}{J.~Duan},
\newblock \bibinfo{title}{Learning the temporal evolution of multivariate
  densities via normalizing flows},
\newblock \bibinfo{journal}{Chaos: An Interdisciplinary Journal of Nonlinear
  Science} \bibinfo{volume}{32} (\bibinfo{year}{2022}).
\bibitem[{Dasgupta et~al.(2024)Dasgupta, Patel, Ray, Johnson, and
  Oberai}]{dasgupta2024dimension}
\bibinfo{author}{A.~Dasgupta}, \bibinfo{author}{D.~V. Patel},
  \bibinfo{author}{D.~Ray}, \bibinfo{author}{E.~A. Johnson},
  \bibinfo{author}{A.~A. Oberai},
\newblock \bibinfo{title}{A dimension-reduced variational approach for solving
  physics-based inverse problems using generative adversarial network priors
  and normalizing flows},
\newblock \bibinfo{journal}{Computer Methods in Applied Mechanics and
  Engineering} \bibinfo{volume}{420} (\bibinfo{year}{2024})
  \bibinfo{pages}{116682}.
\bibitem[{Gutmann and Hyv{\"a}rinen(2010)}]{gutmann2010noise}
\bibinfo{author}{M.~Gutmann}, \bibinfo{author}{A.~Hyv{\"a}rinen},
\newblock \bibinfo{title}{Noise-contrastive estimation: A new estimation
  principle for unnormalized statistical models},
\newblock in: \bibinfo{booktitle}{Proceedings of the Thirteenth International
  Conference on Artificial Intelligence and Statistics},
  \bibinfo{organization}{JMLR Workshop and Conference Proceedings},
  \bibinfo{year}{2010}, pp. \bibinfo{pages}{297--304}.
\bibitem[{Rhodes et~al.(2020)Rhodes, Xu, and Gutmann}]{rhodes2020telescoping}
\bibinfo{author}{B.~Rhodes}, \bibinfo{author}{K.~Xu}, \bibinfo{author}{M.~U.
  Gutmann},
\newblock \bibinfo{title}{Telescoping density-ratio estimation},
\newblock \bibinfo{journal}{Advances in Neural Information Processing Systems}
  \bibinfo{volume}{33} (\bibinfo{year}{2020}) \bibinfo{pages}{4905--4916}.
\bibitem[{Tsimpos et~al.(2025)Tsimpos, Ren, Zech, and
  Marzouk}]{tsimpos2025optimal}
\bibinfo{author}{P.~Tsimpos}, \bibinfo{author}{Z.~Ren},
  \bibinfo{author}{J.~Zech}, \bibinfo{author}{Y.~Marzouk},
\newblock \bibinfo{title}{Optimal scheduling of dynamic transport},
\newblock \bibinfo{journal}{arXiv preprint arXiv:2504.14425}
  (\bibinfo{year}{2025}).
\bibitem[{Lipman et~al.(2022)Lipman, Chen, Ben-Hamu, Nickel, and
  Le}]{lipman2022flow}
\bibinfo{author}{Y.~Lipman}, \bibinfo{author}{R.~T. Chen},
  \bibinfo{author}{H.~Ben-Hamu}, \bibinfo{author}{M.~Nickel},
  \bibinfo{author}{M.~Le},
\newblock \bibinfo{title}{Flow matching for generative modeling},
\newblock \bibinfo{journal}{arXiv preprint arXiv:2210.02747}
  (\bibinfo{year}{2022}).
\bibitem[{Chen et~al.(2018)Chen, Rubanova, Bettencourt, and
  Duvenaud}]{chen2018neural}
\bibinfo{author}{R.~T. Chen}, \bibinfo{author}{Y.~Rubanova},
  \bibinfo{author}{J.~Bettencourt}, \bibinfo{author}{D.~K. Duvenaud},
\newblock \bibinfo{title}{Neural ordinary differential equations},
\newblock \bibinfo{journal}{Advances in neural information processing systems}
  \bibinfo{volume}{31} (\bibinfo{year}{2018}).
\bibitem[{Grathwohl et~al.(2018)Grathwohl, Chen, Bettencourt, Sutskever, and
  Duvenaud}]{grathwohl2018ffjord}
\bibinfo{author}{W.~Grathwohl}, \bibinfo{author}{R.~T. Chen},
  \bibinfo{author}{J.~Bettencourt}, \bibinfo{author}{I.~Sutskever},
  \bibinfo{author}{D.~Duvenaud},
\newblock \bibinfo{title}{{FFJORD}: {F}ree-form continuous dynamics for
  scalable reversible generative models},
\newblock \bibinfo{journal}{arXiv preprint arXiv:1810.01367}
  (\bibinfo{year}{2018}).
\bibitem[{Onken et~al.(2021)Onken, Fung, Li, and Ruthotto}]{onken2021ot}
\bibinfo{author}{D.~Onken}, \bibinfo{author}{S.~W. Fung},
  \bibinfo{author}{X.~Li}, \bibinfo{author}{L.~Ruthotto},
\newblock \bibinfo{title}{{OT-Flow}: {F}ast and accurate continuous normalizing
  flows via optimal transport},
\newblock in: \bibinfo{booktitle}{Proceedings of the AAAI Conference on
  Artificial Intelligence}, volume~\bibinfo{volume}{35}, \bibinfo{year}{2021},
  pp. \bibinfo{pages}{9223--9232}.
\bibitem[{Su et~al.(2022)Su, Song, Meng, and Ermon}]{su2022dual}
\bibinfo{author}{X.~Su}, \bibinfo{author}{J.~Song}, \bibinfo{author}{C.~Meng},
  \bibinfo{author}{S.~Ermon},
\newblock \bibinfo{title}{Dual diffusion implicit bridges for image-to-image
  translation},
\newblock \bibinfo{journal}{arXiv preprint arXiv:2203.08382}
  (\bibinfo{year}{2022}).
\bibitem[{Albergo and Vanden-Eijnden(2022)}]{albergo2022building}
\bibinfo{author}{M.~S. Albergo}, \bibinfo{author}{E.~Vanden-Eijnden},
\newblock \bibinfo{title}{Building normalizing flows with stochastic
  interpolants},
\newblock \bibinfo{journal}{arXiv preprint arXiv:2209.15571}
  (\bibinfo{year}{2022}).
\bibitem[{Albergo et~al.(2023)Albergo, Boffi, and
  Vanden-Eijnden}]{albergo2023stochastic}
\bibinfo{author}{M.~S. Albergo}, \bibinfo{author}{N.~M. Boffi},
  \bibinfo{author}{E.~Vanden-Eijnden},
\newblock \bibinfo{title}{Stochastic interpolants: A unifying framework for
  flows and diffusions},
\newblock \bibinfo{journal}{arXiv preprint arXiv:2303.08797}
  (\bibinfo{year}{2023}).
\bibitem[{Liu et~al.(2022)Liu, Gong, and Liu}]{liu2022flow}
\bibinfo{author}{X.~Liu}, \bibinfo{author}{C.~Gong}, \bibinfo{author}{Q.~Liu},
\newblock \bibinfo{title}{Flow straight and fast: {L}earning to generate and
  transfer data with rectified flow},
\newblock \bibinfo{journal}{arXiv preprint arXiv:2209.03003}
  (\bibinfo{year}{2022}).
\bibitem[{Liu(2022)}]{liu2022rectified}
\bibinfo{author}{Q.~Liu},
\newblock \bibinfo{title}{Rectified flow: A marginal preserving approach to
  optimal transport},
\newblock \bibinfo{journal}{arXiv preprint arXiv:2209.14577}
  (\bibinfo{year}{2022}).
\bibitem[{Pavon et~al.(2021)Pavon, Trigila, and Tabak}]{pavon2021data}
\bibinfo{author}{M.~Pavon}, \bibinfo{author}{G.~Trigila},
  \bibinfo{author}{E.~G. Tabak},
\newblock \bibinfo{title}{The data-driven {S}chr{\"o}dinger {B}ridge},
\newblock \bibinfo{journal}{Communications on Pure and Applied Mathematics}
  \bibinfo{volume}{74} (\bibinfo{year}{2021}) \bibinfo{pages}{1545--1573}.
\bibitem[{De~Bortoli et~al.(2021)De~Bortoli, Thornton, Heng, and
  Doucet}]{de2021diffusion}
\bibinfo{author}{V.~De~Bortoli}, \bibinfo{author}{J.~Thornton},
  \bibinfo{author}{J.~Heng}, \bibinfo{author}{A.~Doucet},
\newblock \bibinfo{title}{Diffusion {S}chr{\"o}dinger {B}ridge with
  applications to score-based generative modeling},
\newblock \bibinfo{journal}{Advances in Neural Information Processing Systems}
  \bibinfo{volume}{34} (\bibinfo{year}{2021}) \bibinfo{pages}{17695--17709}.
\bibitem[{Chen et~al.(2021)Chen, Liu, and Theodorou}]{chen2021likelihood}
\bibinfo{author}{T.~Chen}, \bibinfo{author}{G.-H. Liu}, \bibinfo{author}{E.~A.
  Theodorou},
\newblock \bibinfo{title}{Likelihood training of {S}chr{\"o}dinger bridge using
  forward-backward sdes theory},
\newblock \bibinfo{journal}{arXiv preprint arXiv:2110.11291}
  (\bibinfo{year}{2021}).
\bibitem[{Pooladian and Niles-Weed(2024)}]{pooladian2024plug}
\bibinfo{author}{A.-A. Pooladian}, \bibinfo{author}{J.~Niles-Weed},
\newblock \bibinfo{title}{Plug-in estimation of schr$\backslash$" odinger
  {B}ridges},
\newblock \bibinfo{journal}{arXiv preprint arXiv:2408.11686}
  (\bibinfo{year}{2024}).
\bibitem[{Duvenaud et~al.(2020)Duvenaud, Wang, Jacobsen, Swersky, Norouzi, and
  Grathwohl}]{duvenaud2020your}
\bibinfo{author}{D.~Duvenaud}, \bibinfo{author}{J.~Wang},
  \bibinfo{author}{J.~Jacobsen}, \bibinfo{author}{K.~Swersky},
  \bibinfo{author}{M.~Norouzi}, \bibinfo{author}{W.~Grathwohl},
\newblock \bibinfo{title}{Your classifier is secretly an energy based model and
  you should treat it like one},
\newblock in: \bibinfo{booktitle}{International Conference on Learning
  Representations}, \bibinfo{year}{2020}.
\bibitem[{Ferrer(2022)}]{ferrer2022analysis}
\bibinfo{author}{L.~Ferrer},
\newblock \bibinfo{title}{Analysis and comparison of classification metrics},
\newblock \bibinfo{journal}{arXiv preprint arXiv:2209.05355}
  (\bibinfo{year}{2022}).
\bibitem[{Ruder(2016)}]{ruder2016overview}
\bibinfo{author}{S.~Ruder},
\newblock \bibinfo{title}{An overview of gradient descent optimization
  algorithms},
\newblock \bibinfo{journal}{arXiv preprint arXiv:1609.04747}
  (\bibinfo{year}{2016}).
\bibitem[{Soize and Ghanem(2016)}]{soize2016data}
\bibinfo{author}{C.~Soize}, \bibinfo{author}{R.~Ghanem},
\newblock \bibinfo{title}{Data-driven probability concentration and sampling on
  manifold},
\newblock \bibinfo{journal}{Journal of Computational Physics}
  \bibinfo{volume}{321} (\bibinfo{year}{2016}) \bibinfo{pages}{242--258}.
\bibitem[{Kapusniak et~al.(2024)Kapusniak, Potaptchik, Reu, Zhang, Tong,
  Bronstein, Bose, and Di~Giovanni}]{kapusniak2024metric}
\bibinfo{author}{K.~Kapusniak}, \bibinfo{author}{P.~Potaptchik},
  \bibinfo{author}{T.~Reu}, \bibinfo{author}{L.~Zhang},
  \bibinfo{author}{A.~Tong}, \bibinfo{author}{M.~Bronstein},
  \bibinfo{author}{J.~Bose}, \bibinfo{author}{F.~Di~Giovanni},
\newblock \bibinfo{title}{Metric flow matching for smooth interpolations on the
  data manifold},
\newblock \bibinfo{journal}{Advances in Neural Information Processing Systems}
  \bibinfo{volume}{37} (\bibinfo{year}{2024}) \bibinfo{pages}{135011--135042}.
\bibitem[{Baydin et~al.(2018)Baydin, Pearlmutter, Radul, and
  Siskind}]{baydin2018automatic}
\bibinfo{author}{A.~G. Baydin}, \bibinfo{author}{B.~A. Pearlmutter},
  \bibinfo{author}{A.~A. Radul}, \bibinfo{author}{J.~M. Siskind},
\newblock \bibinfo{title}{Automatic differentiation in machine learning: {A}
  survey},
\newblock \bibinfo{journal}{Journal of machine learning research}
  \bibinfo{volume}{18} (\bibinfo{year}{2018}) \bibinfo{pages}{1--43}.
\bibitem[{Girolami and Calderhead(2011)}]{girolami2011riemann}
\bibinfo{author}{M.~Girolami}, \bibinfo{author}{B.~Calderhead},
\newblock \bibinfo{title}{Riemann {M}anifold {L}angevin and {H}amiltonian
  {M}onte {C}arlo methods},
\newblock \bibinfo{journal}{Journal of the Royal Statistical Society Series
  {B}: Statistical Methodology} \bibinfo{volume}{73} (\bibinfo{year}{2011})
  \bibinfo{pages}{123--214}.
\bibitem[{Gottwald et~al.(2024)Gottwald, Li, Marzouk, and
  Reich}]{gottwald2024stable}
\bibinfo{author}{G.~A. Gottwald}, \bibinfo{author}{F.~Li},
  \bibinfo{author}{Y.~Marzouk}, \bibinfo{author}{S.~Reich},
\newblock \bibinfo{title}{Stable generative modeling using schr\"{o}dinger
  {B}ridges},
\newblock \bibinfo{journal}{arXiv preprint arXiv:2401.04372}
  (\bibinfo{year}{2024}).
\bibitem[{Goodfellow et~al.(2020)Goodfellow, Pouget-Abadie, Mirza, Xu,
  Warde-Farley, Ozair, Courville, and Bengio}]{goodfellow2020generative}
\bibinfo{author}{I.~Goodfellow}, \bibinfo{author}{J.~Pouget-Abadie},
  \bibinfo{author}{M.~Mirza}, \bibinfo{author}{B.~Xu},
  \bibinfo{author}{D.~Warde-Farley}, \bibinfo{author}{S.~Ozair},
  \bibinfo{author}{A.~Courville}, \bibinfo{author}{Y.~Bengio},
\newblock \bibinfo{title}{Generative {A}dversarial {N}etworks},
\newblock \bibinfo{journal}{Communications of the ACM} \bibinfo{volume}{63}
  (\bibinfo{year}{2020}) \bibinfo{pages}{139--144}.
\bibitem[{Kobyzev et~al.(2020)Kobyzev, Prince, and
  Brubaker}]{kobyzev2020normalizing}
\bibinfo{author}{I.~Kobyzev}, \bibinfo{author}{S.~J. Prince},
  \bibinfo{author}{M.~A. Brubaker},
\newblock \bibinfo{title}{Normalizing flows: {A}n introduction and review of
  current methods},
\newblock \bibinfo{journal}{IEEE transactions on pattern analysis and machine
  intelligence} \bibinfo{volume}{43} (\bibinfo{year}{2020})
  \bibinfo{pages}{3964--3979}.
\bibitem[{Lipman et~al.(2024)Lipman, Havasi, Holderrieth, Shaul, Le, Karrer,
  Chen, Lopez-Paz, Ben-Hamu, and Gat}]{lipman2024flow}
\bibinfo{author}{Y.~Lipman}, \bibinfo{author}{M.~Havasi},
  \bibinfo{author}{P.~Holderrieth}, \bibinfo{author}{N.~Shaul},
  \bibinfo{author}{M.~Le}, \bibinfo{author}{B.~Karrer}, \bibinfo{author}{R.~T.
  Chen}, \bibinfo{author}{D.~Lopez-Paz}, \bibinfo{author}{H.~Ben-Hamu},
  \bibinfo{author}{I.~Gat},
\newblock \bibinfo{title}{Flow matching guide and code},
\newblock \bibinfo{journal}{arXiv preprint arXiv:2412.06264}
  (\bibinfo{year}{2024}).
\bibitem[{Song et~al.(2020)Song, Sohl-Dickstein, Kingma, Kumar, Ermon, and
  Poole}]{song2020score}
\bibinfo{author}{Y.~Song}, \bibinfo{author}{J.~Sohl-Dickstein},
  \bibinfo{author}{D.~P. Kingma}, \bibinfo{author}{A.~Kumar},
  \bibinfo{author}{S.~Ermon}, \bibinfo{author}{B.~Poole},
\newblock \bibinfo{title}{Score-based generative modeling through stochastic
  differential equations},
\newblock \bibinfo{journal}{arXiv preprint arXiv:2011.13456}
  (\bibinfo{year}{2020}).
\bibitem[{Dasgupta et~al.(2025)Dasgupta, da~Cunha, Fardisi, Aminy, Binder,
  Shaddy, and Oberai}]{dasgupta2025unifying}
\bibinfo{author}{A.~Dasgupta}, \bibinfo{author}{A.~M. da~Cunha},
  \bibinfo{author}{A.~Fardisi}, \bibinfo{author}{M.~Aminy},
  \bibinfo{author}{B.~Binder}, \bibinfo{author}{B.~Shaddy},
  \bibinfo{author}{A.~A. Oberai},
\newblock \bibinfo{title}{Unifying and extending diffusion models through
  {PDE}s for solving inverse problems},
\newblock \bibinfo{journal}{arXiv preprint arXiv:2504.07437}
  (\bibinfo{year}{2025}).
\bibitem[{Song and Kingma(2021)}]{song2021train}
\bibinfo{author}{Y.~Song}, \bibinfo{author}{D.~P. Kingma},
\newblock \bibinfo{title}{How to train your energy-based models},
\newblock \bibinfo{journal}{arXiv preprint arXiv:2101.03288}
  (\bibinfo{year}{2021}).
\bibitem[{Du et~al.(2020)Du, Li, Tenenbaum, and Mordatch}]{du2020improved}
\bibinfo{author}{Y.~Du}, \bibinfo{author}{S.~Li},
  \bibinfo{author}{J.~Tenenbaum}, \bibinfo{author}{I.~Mordatch},
\newblock \bibinfo{title}{Improved contrastive divergence training of
  energy-based models},
\newblock \bibinfo{journal}{arXiv preprint arXiv:2012.01316}
  (\bibinfo{year}{2020}).
\bibitem[{Gao et~al.(2020)Gao, Nijkamp, Kingma, Xu, Dai, and Wu}]{gao2020flow}
\bibinfo{author}{R.~Gao}, \bibinfo{author}{E.~Nijkamp}, \bibinfo{author}{D.~P.
  Kingma}, \bibinfo{author}{Z.~Xu}, \bibinfo{author}{A.~M. Dai},
  \bibinfo{author}{Y.~N. Wu},
\newblock \bibinfo{title}{Flow contrastive estimation of energy-based models},
\newblock in: \bibinfo{booktitle}{Proceedings of the IEEE/CVF Conference on
  Computer Vision and Pattern Recognition}, \bibinfo{year}{2020}, pp.
  \bibinfo{pages}{7518--7528}.
\bibitem[{Tabandeh et~al.(2022)Tabandeh, Sharma, Iannacone, and
  Gardoni}]{tabandeh2022numerical}
\bibinfo{author}{A.~Tabandeh}, \bibinfo{author}{N.~Sharma},
  \bibinfo{author}{L.~Iannacone}, \bibinfo{author}{P.~Gardoni},
\newblock \bibinfo{title}{Numerical solution of the {F}okker--{P}lanck equation
  using physics-based mixture models},
\newblock \bibinfo{journal}{Computer Methods in Applied Mechanics and
  Engineering} \bibinfo{volume}{399} (\bibinfo{year}{2022})
  \bibinfo{pages}{115424}.
\bibitem[{Ueda(1979)}]{ueda1979randomly}
\bibinfo{author}{Y.~Ueda},
\newblock \bibinfo{title}{Randomly transitional phenomena in the system
  governed by duffing's equation},
\newblock \bibinfo{journal}{Journal of Statistical Physics}
  \bibinfo{volume}{20} (\bibinfo{year}{1979}) \bibinfo{pages}{181--196}.
\bibitem[{Hammad et~al.(2020)Hammad, Salas, and El-Tantawy}]{hammad2020new}
\bibinfo{author}{M.~A. Hammad}, \bibinfo{author}{A.~H. Salas},
  \bibinfo{author}{S.~El-Tantawy},
\newblock \bibinfo{title}{New method for solving strong conservative odd parity
  nonlinear oscillators: applications to plasma physics and rigid rotator},
\newblock \bibinfo{journal}{AIP Advances} \bibinfo{volume}{10}
  (\bibinfo{year}{2020}).
\bibitem[{Silverman(2018)}]{silverman2018density}
\bibinfo{author}{B.~W. Silverman}, \bibinfo{title}{Density estimation for
  statistics and data analysis}, \bibinfo{publisher}{Routledge},
  \bibinfo{year}{2018}.
\bibitem[{Ismail et~al.(2009)Ismail, Ikhouane, and
  Rodellar}]{ismail2009hysteresis}
\bibinfo{author}{M.~Ismail}, \bibinfo{author}{F.~Ikhouane},
  \bibinfo{author}{J.~Rodellar},
\newblock \bibinfo{title}{The hysteresis {B}ouc-{W}en model: {A} survey},
\newblock \bibinfo{journal}{Archives of computational methods in engineering}
  \bibinfo{volume}{16} (\bibinfo{year}{2009}) \bibinfo{pages}{161--188}.
\bibitem[{Wen(1976)}]{wen1976method}
\bibinfo{author}{Y.-K. Wen},
\newblock \bibinfo{title}{Method for random vibration of hysteretic systems},
\newblock \bibinfo{journal}{Journal of the engineering mechanics division}
  \bibinfo{volume}{102} (\bibinfo{year}{1976}) \bibinfo{pages}{249--263}.
\bibitem[{Wen(1980)}]{wen1980equivalent}
\bibinfo{author}{Y.~Wen},
\newblock \bibinfo{title}{Equivalent linearization for hysteretic systems under
  random excitation}  (\bibinfo{year}{1980}).
\bibitem[{Cuturi(2013)}]{cuturi2013sinkhorn}
\bibinfo{author}{M.~Cuturi},
\newblock \bibinfo{title}{Sinkhorn distances: Lightspeed computation of optimal
  transport},
\newblock \bibinfo{journal}{Advances in neural information processing systems}
  \bibinfo{volume}{26} (\bibinfo{year}{2013}).
\bibitem[{Hoffman and Gelman(2014)}]{hoffman2014no}
\bibinfo{author}{M.~D. Hoffman}, \bibinfo{author}{A.~Gelman},
\newblock \bibinfo{title}{The {No-U-Turn} sampler: {A}daptively setting path
  lengths in {H}amiltonian {M}onte {C}arlo},
\newblock \bibinfo{journal}{J. Mach. Learn. Res.} \bibinfo{volume}{15}
  (\bibinfo{year}{2014}) \bibinfo{pages}{1593--1623}.
\bibitem[{Cobb(2023)}]{cobb2023hamiltorch}
\bibinfo{author}{A.~D. Cobb},
\newblock \bibinfo{title}{\texttt{hamiltorch}: {A} {PyTorch}-based library for
  {H}amiltonian {M}onte {C}arlo},
\newblock in: \bibinfo{booktitle}{Proceedings of Cyber-Physical Systems and
  Internet of Things}, \bibinfo{year}{2023}, pp. \bibinfo{pages}{114--115}.
\bibitem[{Han et~al.(2022)Han, Hu, Huang, Jiang, and Zhao}]{han2022adbench}
\bibinfo{author}{S.~Han}, \bibinfo{author}{X.~Hu}, \bibinfo{author}{H.~Huang},
  \bibinfo{author}{M.~Jiang}, \bibinfo{author}{Y.~Zhao},
\newblock \bibinfo{title}{{ADBench}: {A}nomaly {D}etection {B}enchmark},
\newblock \bibinfo{journal}{Advances in neural information processing systems}
  \bibinfo{volume}{35} (\bibinfo{year}{2022}) \bibinfo{pages}{32142--32159}.
\bibitem[{Krizhevsky et~al.(2009)Krizhevsky, Hinton
  et~al.}]{krizhevsky2009learning}
\bibinfo{author}{A.~Krizhevsky}, \bibinfo{author}{G.~Hinton}, et~al.,
  \bibinfo{title}{Learning multiple layers of features from tiny images},
  \bibinfo{type}{Technical Report}, Toronto, ON, Canada, \bibinfo{year}{2009}.
\bibitem[{Xiao et~al.(2017)Xiao, Rasul, and Vollgraf}]{xiao2017}
\bibinfo{author}{H.~Xiao}, \bibinfo{author}{K.~Rasul},
  \bibinfo{author}{R.~Vollgraf},
\newblock \bibinfo{title}{{Fashion-MNIST}: {A} novel image dataset for
  benchmarking machine learning algorithms},
\newblock \bibinfo{journal}{arXiv preprint arXiv:1708.07747}
  (\bibinfo{year}{2017}).
\bibitem[{Mu and Gilmer(2019)}]{mu2019mnist}
\bibinfo{author}{N.~Mu}, \bibinfo{author}{J.~Gilmer},
\newblock \bibinfo{title}{{MNIST-C}: {A} robustness benchmark for computer
  vision},
\newblock \bibinfo{journal}{arXiv preprint arXiv:1906.02337}
  (\bibinfo{year}{2019}).
\bibitem[{Kim et~al.(2023)Kim, Hwang, Lee, Kim, and Kim}]{kim2023odim}
\bibinfo{author}{D.~Kim}, \bibinfo{author}{J.~Hwang}, \bibinfo{author}{J.~Lee},
  \bibinfo{author}{K.~Kim}, \bibinfo{author}{Y.~Kim},
\newblock \bibinfo{title}{{ODIM}: {O}utlier detection via likelihood of
  under-fitted generative models},
\newblock \bibinfo{journal}{arXiv preprint arXiv:2301.04257}
  (\bibinfo{year}{2023}).
\bibitem[{Tancik et~al.(2020)Tancik, Srinivasan, Mildenhall, Fridovich-Keil,
  Raghavan, Singhal, Ramamoorthi, Barron, and Ng}]{tancik2020fourier}
\bibinfo{author}{M.~Tancik}, \bibinfo{author}{P.~Srinivasan},
  \bibinfo{author}{B.~Mildenhall}, \bibinfo{author}{S.~Fridovich-Keil},
  \bibinfo{author}{N.~Raghavan}, \bibinfo{author}{U.~Singhal},
  \bibinfo{author}{R.~Ramamoorthi}, \bibinfo{author}{J.~Barron},
  \bibinfo{author}{R.~Ng},
\newblock \bibinfo{title}{Fourier features let networks learn high frequency
  functions in low dimensional domains},
\newblock \bibinfo{journal}{Advances in neural information processing systems}
  \bibinfo{volume}{33} (\bibinfo{year}{2020}) \bibinfo{pages}{7537--7547}.
\bibitem[{Baptista et~al.(2025)Baptista, Dasgupta, Kovachki, Oberai, and
  Stuart}]{baptista2025memorization}
\bibinfo{author}{R.~Baptista}, \bibinfo{author}{A.~Dasgupta},
  \bibinfo{author}{N.~B. Kovachki}, \bibinfo{author}{A.~Oberai},
  \bibinfo{author}{A.~M. Stuart},
\newblock \bibinfo{title}{Memorization and regularization in generative
  diffusion models},
\newblock \bibinfo{journal}{arXiv preprint arXiv:2501.15785}
  (\bibinfo{year}{2025}).
\bibitem[{Virtanen et~al.(2020)Virtanen, Gommers, Oliphant, Haberland, Reddy,
  Cournapeau, Burovski, Peterson, Weckesser, Bright, {van der Walt}, Brett,
  Wilson, Millman, Mayorov, Nelson, Jones, Kern, Larson, Carey, Polat, Feng,
  Moore, {VanderPlas}, Laxalde, Perktold, Cimrman, Henriksen, Quintero, Harris,
  Archibald, Ribeiro, Pedregosa, {van Mulbregt}, and {SciPy 1.0
  Contributors}}]{2020SciPy-NMeth}
\bibinfo{author}{P.~Virtanen}, \bibinfo{author}{R.~Gommers},
  \bibinfo{author}{T.~E. Oliphant}, \bibinfo{author}{M.~Haberland},
  \bibinfo{author}{T.~Reddy}, \bibinfo{author}{D.~Cournapeau},
  \bibinfo{author}{E.~Burovski}, \bibinfo{author}{P.~Peterson},
  \bibinfo{author}{W.~Weckesser}, \bibinfo{author}{J.~Bright},
  \bibinfo{author}{S.~J. {van der Walt}}, \bibinfo{author}{M.~Brett},
  \bibinfo{author}{J.~Wilson}, \bibinfo{author}{K.~J. Millman},
  \bibinfo{author}{N.~Mayorov}, \bibinfo{author}{A.~R.~J. Nelson},
  \bibinfo{author}{E.~Jones}, \bibinfo{author}{R.~Kern},
  \bibinfo{author}{E.~Larson}, \bibinfo{author}{C.~J. Carey},
  \bibinfo{author}{{\.I}.~Polat}, \bibinfo{author}{Y.~Feng},
  \bibinfo{author}{E.~W. Moore}, \bibinfo{author}{J.~{VanderPlas}},
  \bibinfo{author}{D.~Laxalde}, \bibinfo{author}{J.~Perktold},
  \bibinfo{author}{R.~Cimrman}, \bibinfo{author}{I.~Henriksen},
  \bibinfo{author}{E.~A. Quintero}, \bibinfo{author}{C.~R. Harris},
  \bibinfo{author}{A.~M. Archibald}, \bibinfo{author}{A.~H. Ribeiro},
  \bibinfo{author}{F.~Pedregosa}, \bibinfo{author}{P.~{van Mulbregt}},
  \bibinfo{author}{{SciPy 1.0 Contributors}},
\newblock \bibinfo{title}{{{SciPy} 1.0: {F}undamental {A}lgorithms for
  {S}cientific {C}omputing in {P}ython}},
\newblock \bibinfo{journal}{Nature Methods} \bibinfo{volume}{17}
  (\bibinfo{year}{2020}) \bibinfo{pages}{261--272}.
  \DOIprefix\doi{10.1038/s41592-019-0686-2}.

\end{thebibliography}
	
\end{document}